\documentclass[10pt,twocolumn,letterpaper]{article}

\usepackage[pagenumbers]{cvpr} % To force page numbers, e.g. for an arXiv version

\usepackage{makecell}

\definecolor{cvprblue}{rgb}{0.21,0.49,0.74}
\usepackage[pagebackref,breaklinks,colorlinks,allcolors=cvprblue]{hyperref}

\def\paperID{*****} % *** Enter the Paper ID here
\def\confName{}
\def\confYear{2026}

\title{See-through: Single-image Layer Decomposition for Anime Characters}

\author{
    Jian Lin$^1$ \quad 
    Chengze Li$^{1}$\thanks{Corresponding author.} \quad 
    Haoyun Qin$^{2,3,4}$ \quad 
    Kwun Wang Chan$^1$ \\
    Yanghua Jin$^3$ \quad 
    Hanyuan Liu$^1$ \quad 
    Chun Wang Stephen Choy$^1$ \quad 
    Xueting Liu$^1$ \\[2mm]
    $^1$Saint Francis University \quad $^2$University of Pennsylvania \quad $^3$Spellbrush \quad $^4$Shitagaki Lab \\
    {\tt\small \{jlin, czli$^{*}$, tliu\}@sfu.edu.hk, qhy@seas.upenn.edu,} \\
    {\tt\small chankwunwang68@gmail.com, yanghua@spellbrush.com, lkwq007@gmail.com, stephenchoy626@gmail.com}
}

\cvprteaser{
    \centering
    \includegraphics[width=\linewidth]{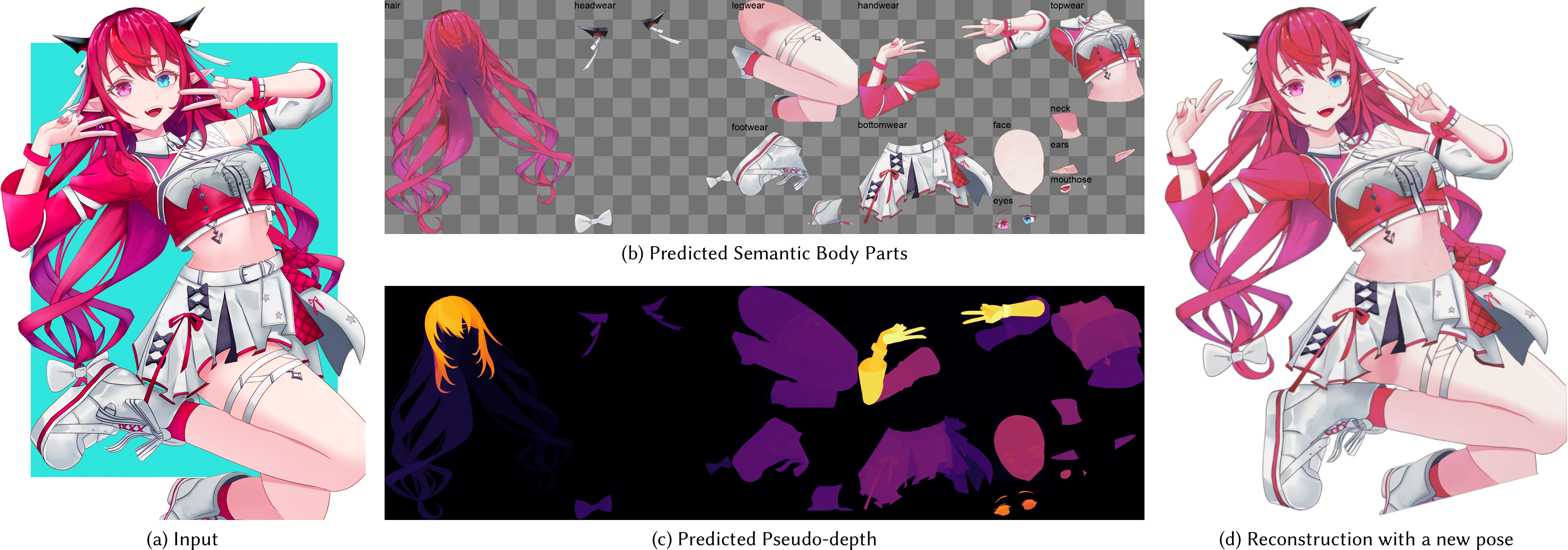}
    \caption{\textbf{From a single illustration to a 2.5D-ready character.} Given an input anime image (a), our method generates fully inpainted, semantic RGBA body-part layers (b) and per-part pseudo-depth for stratified ordering (c), enabling faithful re-rendering and reposing in a 2.5D pipeline (d). \textcopyright Stephen Choy.}
    \label{fig:teaser}
}

\begin{document}
\maketitle

\begin{abstract}
We introduce a framework that automates the transformation of static anime illustrations into manipulatable 2.5D models. Current professional workflows require tedious manual segmentation and the artistic ``hallucination'' of occluded regions to enable motion. Our approach overcomes this by decomposing a single image into fully inpainted, semantically distinct layers with inferred drawing orders. To address the scarcity of training data, we introduce a scalable engine that bootstraps high-quality supervision from commercial Live2D models, capturing pixel-perfect semantics and hidden geometry. Our methodology couples a diffusion-based Body Part Consistency Module, which enforces global geometric coherence, with a pixel-level pseudo-depth inference mechanism. This combination resolves the intricate stratification of anime characters, e.g., interleaving hair strands, allowing for dynamic layer reconstruction. We demonstrate that our approach yields high-fidelity, manipulatable models suitable for professional, real-time animation applications.
\end{abstract}

\section{Introduction}

The anime community actively produces a wealth of content to entertain global audiences. A surging trend involves transforming high-quality static illustrations (\textit{Tachi-e}) into interactive ``motion illustrations,'' which serve as the visual core for VTubing, 2D games, and visual novels. However, realising this transformation is non-trivial; the challenge lies in providing high-quality animation and interactivity while strictly preserving the original aesthetic integrity, which makes a full 3D modelling pipeline often impractical and aesthetically incompatible for anime artwork.

To resolve this, the industry typically adopts a ``2.5D'' approach, such as \textit{Live2D}. In this workflow, artists extend the original flat illustration into semantically meaningful 2D layers and draw additional visual content to cover previously occluded regions required for motion (as in Figure~\ref{fig:teaser}(b)). These parts are then composited within the rendering engine based on artist-specified drawing order. By manipulating these layers with motion and deformation timings, creators can achieve visually pleasing effects, which creates the illusion of 3D volume while strictly maintaining the original 2D aesthetics (Figure~\ref{fig:teaser}(d)).
Despite its visual advantages, this workflow imposes a massive workload on creators. They must manually separate a single character illustration into dozens or hundreds of layers and explicitly determine their precise placement order. Crucially, they have to hallucinate the occluded regions for all layers with plausible visuals, which is tedious and technically demanding.

To the best of our knowledge, there is currently no direct solution to address this workflow. While recent generative approaches offer consistent transparent layer decomposition~\cite{layerdiffusion, qwen-image-layered} or scene object deocclusion~\cite{object-level-scene-deocclusion}, they struggle to provide a comprehensive, fine-grained collection of layers required for detailed animation. Most critically, these models typically assume a fixed layer ordering, failing to account for the complex stratification inherent to anime characters. We demonstrate this case in Figure~\ref{fig:teaser}, where the single layer of hair is visually wrapping over the face-related layers.

In this work, we present a novel framework that automates the conversion of a single static illustration into a fully separated, 2.5D-ready character model. Our approach decomposes the image into 4-channel, inpainted, semantically distinct body part layers and infers a fine-grained drawing order for them.
Because no ground-truth annotations exist for 2.5D-ready anime models, we build a robust data engine that bootstraps labels from 2D semantic segmentation pipelines. The data engine leverages the weak supervision of GradCAM~\cite{GradCAMPlusPlus} and the segmentation prior of SAM~\cite{sam} to gather rough 2D segmentations, and then propagate them into 2.5D labels with the help of the Live2D rendering engine. This process yields pixel-perfect supervision for 19 semantic body parts, their occluded regions, and fragment-level drawing order, providing a practical ``see-through'' training signal for reconstructing hidden anatomy.
Leveraging this dataset, we develop our body decomposition framework based on Latent Diffusion Models~\cite{sdxl}. We construct the training of the framework with a two-stage strategy: the first stage learns high-fidelity synthesis of individual body part RGBA layers, and the second stage introduces a Body Part Consistency Module to refine all parts jointly for completeness and cross-layer coherence. We also apply this strategy for inferring a fine-grained drawing order in the form of pseudo-depth, which supports complex stratification, including interleaving within a single semantic layer.

We evaluate the efficacy of our framework through experiments that demonstrate that it achieves precise and aesthetically faithful reconstructions suitable for production-level 2.5D animation. Our model also serves as a robust tool for fine-grained 2D anime body parsing. We validate our performance through qualitative and quantitative evaluations.
To illustrate practical versatility, we also present several real-time animation applications, which benefit substantially from our decomposed and stratified layers.

We summarise our contributions as follows:

\begin{itemize}[leftmargin=2em]
\item We present the first end-to-end framework that converts a \textbf{single} composite anime illustration into a 2.5D-ready character by generating complete semantic RGBA body part layers with occlusion completion and stratified ordering for production.
\item We build a data engine that bootstraps scarce 2.5D labels from weak supervision, producing pixel-perfect annotations for 19 parts, occluded regions, and fragment-level drawing order.
\item We introduce a two-stage latent-diffusion training strategy with a Body Part Consistency Module that improves completeness and cross-layer coherence for our tasks.
\end{itemize}

\begin{figure*}[!t]
    \centering
    \includegraphics[width=\linewidth]{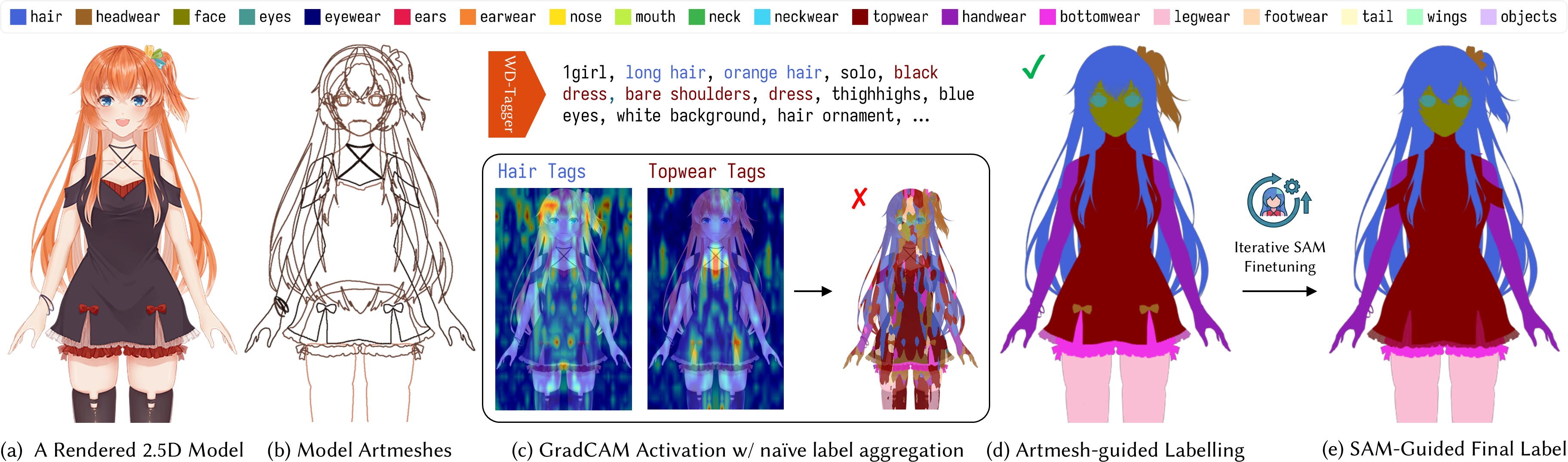}
    \caption{Data engine for 2D anime body part segmentation. We derive coarse ``seed'' masks from Grad-CAM responses of individual classes, and snap them to Live2D ArtMesh visibility masks for pixel-accurate boundaries. We then refine the masks with the SAM prior, producing our final labels. \textcopyright USTC LEO ACG Club.}
    \label{fig:dataset_2d_engine}
\end{figure*}

\section{Related Work}
\label{sec:related_work}

\subsection{Semantic Parsing of Anime Characters}

Precise semantic parsing is a prerequisite for applying animation heuristics to static anime characters. While early solutions relied on interactive user input~\cite{interactiveedgeaware}, the field has shifted towards fully automatic, data-driven architectures. To capture body structure, approaches such as~\cite{bizarre-pose-estimator, cpnet, bodypartsegmentation} leverage keypoint estimation, with the latter two extending these structural priors to achieve dense pixel-level segmentation. Others harness foundation models: \cite{seganimechara} adapts foundation model for a rough anime parsing, while DaCon~\cite{dacon} utilises deep features for semantic region matching. To facilitate these tasks, benchmarks have expanded beyond method-specific datasets to include facial parsing~\cite{animesemanticsegmentationgan, parsingconditioned} and multi-style pose estimation~\cite{humanart}. Finally, while generalist models like the SAM series~\cite{sam, sam3} exhibit strong generalisation, they typically rely on interactive prompting and still have a domain gap towards anime-style imagery.
Crucially, regardless of the methodology, all aforementioned approaches are fundamentally limited to planar 2D parsing. They successfully assign semantics to visible pixels but lack the ``see-through'' capability required to infer the occluded anatomy beneath, which is essential for constructing layered 2.5D animation.

\subsection{Image Layer Decomposition and Z-Order Inference}

Recent diffusion-based methods have enabled the generation and decomposition of 4-channel RGBA layers~\cite{layerdiffusion, layerd, qwen-image-layered}. However, these approaches typically assume a fixed, top-to-bottom layer ordering, which proves insufficient for anime characters where semantic parts exhibit intricate stratification, such as hair strands ``sandwiching'' the face. While amodal completion and scene deocclusion methods~\cite{vinv, object-level-scene-deocclusion} explicitly address stratification, they generally operate at a coarse instance level. They often prioritise individual object completion at the expense of global consistency. Furthermore, they usually infer drawing orders represented in simple directed graphs, which also cannot capture the capture the complex interleaving occlusion relationships.

To represent or resolve layer ordering, traditional graphics approaches have explored local layering~\cite{locallayering} or junction-based heuristics~\cite{25dcartoonhair, stereoscopizingcelanimations}, though these remain largely manual or fragile in complex scenarios. Conversely, modern monocular depth estimators like Depth Anything~\cite{depthanything} and Marigold~\cite{marigold} provide dense pixel-level cues. Although primarily designed for natural photorealistic imagery, these generic depth priors offer a promising avenue for inferring the relative Z-ordering of graphic elements in our domain.

\subsection{Animating Anime Characters}

To animate static drawings, the industry standard relies on warping manually segmented layers with deformations such as affine transformations, or ARAP~\cite{arap}. While early automation in 2.5D cartoon attempts utilised pseudo-3D proxies~\cite{25dcartoonmodels, doublesided25dgraphics} or standardised character sheets~\cite{live2d,generating25dcharacteranimation, viewdependentformulation} for more anime-like animations, these workflows remain heavily dependent on manual layering and annotations. Recent generative approaches aim to automate \textit{Live2D} modelling directly from text~\cite{textoon} or single portraits~\cite{cartoonalive}. However, these methods typically map textures on pre-defined template meshes rather than inferring intrinsic 2.5D anatomy from highly detailed anime input.

Regarding broader articulation techniques, sketch-based motion systems~\cite{toonsynth, animateddrawings, neuralpuppet, charactergan} and facial animators~\cite{talkingheadanime4} provide accessible tools but often fail to handle the intricate textures and complex multi-part occlusions of high-quality anime. Alternatively, lifting 2D illustrations into 3D meshes~\cite{monstermash, panic3d, charactergen} or applying automatic rigging~\cite{photowakeup, drawingspinup, fromriggingtowaving} can enable motion, yet these pipelines often conflict with the characteristic flat shading and non-Euclidean geometric exaggerations of anime. Similarly, video diffusion models~\cite{animatediff, animateanyone, layeranimate, mikudance} offer stylised generative anime motion, but currently suffer from limited resolution and temporal inconsistency to render intricate details, and cannot run in real time. We believe that by providing precise semantics and separated 2.5D parts, our framework offers a foundational representation that could potentially benefit these downstream animation methodologies.

\section{Dataset}
\label{sec:dataset}

To address the layer decomposition and drawing-order inference tasks, a high-quality labelled dataset is essential. Importantly, our objective requires a data representation that captures each individual semantic region of anime body parts, together with their occluded areas and drawing order.
In this section, we explain our proposed scalable data engine, which bootstraps precise labels from a 2.5D rendering engine and weak 2D semantic segmentation supervision.

\subsection{2D Semantic Segmentation Data Engine}
\label{sec:dataset_engine}

\subsubsection{The Live2D Data Structure}
\label{sec:dataset_intro}
We choose Live2D as the source of our 2.5D dataset because it is the de-facto industry standard and abundant models are publicly available. In a typical production workflow, artists first prepare detailed, hierarchically layered drawings in standard authoring tools (e.g., Photoshop or Krita). The Live2D engine then compiles these assets into a 2.5D character model. Concretely, a Live2D model is encoded as a collection of \textit{ArtMeshes} (Figure~\ref{fig:dataset_2d_engine}(b)), where each ArtMesh is a tessellated triangular mesh mapped to a local region of a texture atlas, corresponding to a small drawing fragment such as a hair strand or a ribbon detail. During rendering, the engine maps the original painting onto these meshes as textures, and assigns each ArtMesh a deterministic drawing order index derived from the artist-specified layer stack, providing a discrete z-buffer signal for occlusion resolution. By inspecting the renderer, we can extract the geometry-based visibility mask for each ArtMesh, allowing us to recover pixel-perfect fragment boundaries together with their visibility and blending information. We leverage this precise geometric signal for our subsequent labelling tasks.

\subsubsection{Annotating and Segmenting 2D Anime Body Semantics}
While an implicit ArtMesh hierarchy may help our labelling, we find that it varies across artists and projects, and cannot provide a standardised semantic signal for our subsequent labelling. We therefore adopt a fixed 19-class taxonomy and aim to label all ArtMeshes accordingly. This is challenging at scale, as a single character can contain hundreds of fragments with complex structures (e.g., hair in Figure~\ref{fig:dataset_2d_engine}(b)). To automate labelling, we first seek to bootstrap supervision from 2D segmentation models at the pixel level and then transfer it back to the underlying 2.5D geometry. However, existing segmenters such as SAM~\cite{sam} exhibit a clear domain gap on anime imagery, and to our knowledge no public dataset provides the full-body, fine-grained parsing granularity required by our 19-class setting. We therefore build a weakly supervised initialisation as ``seed'' labels for later fine-tuning.

Specifically, we leverage the semantic vocabulary of the Danbooru tagging system~\cite{danbooru2021} together with the \textit{wd-eva02-large-tagger-v3}~\cite{wd_eva02_large_tagger_v3} classifier. We define a hierarchical mapping from the overly fine-grained visual tag set to our predefined 19 classes~\footnote{e.g., \textit{blonde\_hair}, \textit{ponytail}, and \textit{ahoge} all map to the class \textit{hair}.}.
For each predicted Danbooru visual tag, we compute its GradCAM++~\cite{GradCAMPlusPlus} activation map and aggregate it into its associated semantic class, i.e., one of our predefined 19 classes, as shown in Figure~\ref{fig:dataset_2d_engine}(c), which provides a coarse spatial response indicating where the classifier attends for that concept.
While informative, these responses are intrinsically low-resolution and diffuse, and a naïve pixel-wise max-pooling across classes leads to fragmented and noisy segmentation (Figure~\ref{fig:dataset_2d_engine}(c), right).
Fortunately, Live2D provides exact visibility masks for each visible ArtMesh. We use these masks to rectify the weak 2D signals: for every mesh fragment, we compute spatially averaged activation scores over all 19 classes within its visible region, and assign the fragment the class with the maximum score. This ArtMesh-guided voting removes much of the activation noise and produces boundary-aligned seed labels (Figure~\ref{fig:dataset_2d_engine}(d)).

\subsubsection{Iterative Refinement via Multi-Decoder SAM}

As observed in Figure~\ref{fig:dataset_2d_engine}(d), while GradCAM provides an initial semantic baseline, ambiguities can still remain in visually similar regions, such as misclassifying the ponytail, the shoulder straps, or conflating small ribbon-like decorations. To rectify this, we leverage the strong, generalisable segmentation priors of SAM-HQ~\cite{sam_hq} models.

Technically, as SAM's architecture is inherently designed for zero-shot inference, it lacks a native mechanism for automatic, fixed-class inference. To adapt it for our taxonomy, we replace the original promptable decoder with 19 independent mask decoders, each dedicated to a specific semantic class. By zeroing out the prompt embeddings during training, we compel each decoder to autonomously learn class-specific feature extraction directly from the image embeddings, effectively transforming the model into a closed-set semantic segmenter. Crucially, we implement a self-refining training loop in which the model and its training data iteratively improve each other. Starting from the initial GradCAM labels, the iterative process gradually improves the predictions and uses the newly predicted labels to update the supervision. At each iteration, we apply the same geometric regularization used in the seeding stage to enforce boundary snapping. As in Figure~\ref{fig:dataset_2d_engine}(e), this bootstrapping process corrects initial errors and successfully disambiguates challenging regions. While minor imperfections may still remain, they can be efficiently corrected in the subsequent 2.5D labelling stage, requiring only minimal manual adjustment of a small number of ArtMeshes.

\subsection{2.5D Layering Annotation}
\label{sec:dataset_fragments}

With a robust 2D segmentation model established, we project the predicted pixel-wise labels back onto the constituent ArtMeshes to prepare our final 2.5D dataset. For visible fragments, we assign the semantic class directly based on pixel majorities; however, fully occluded fragments receive no direct supervision. To address this, we employ a heuristic-based propagation strategy that exploits the Live2D structural convention discussed above. Since artists typically organise semantically related ArtMeshes within the same hierarchy (as discussed in Section~\ref{sec:dataset_intro}), we use this structure and connected component analysis of the mesh topology to propagate labels to hidden parts. This process generates a comprehensive initial set of 2.5D annotations and serves as a high-quality baseline for subsequent human verification. We then develop a custom annotation interface with ``see-through'' capabilities, allowing annotators to inspect occluded layers, toggle visibility by semantic class, and quickly correct mislabelled fragments. Please refer to the supplementary material for details of the propagation heuristics and the annotation user interface design.

Once the ArtMesh labelling is verified, we further augment the dataset by applying standard ``angle following'' animations to interpolate the characters through distinct orientations (up, down, lateral, and diagonals). These animations produce non-rigid pose changes, such as head turning or limb articulations. Since the semantic identity of an ArtMesh is invariant to these deformations, this process yields direct labelling of diverse poses without any additional human annotation effort.

After the labelling process, we record the drawing order of the ArtMesh fragments alongside the semantic labels, and formulate it as a \textit{pseudo-depth} metric by normalising the drawing order into a floating-point value. Crucially, we capture this pseudo-depth at the fine-grained ArtMesh level rather than the coarse 19-class semantic level, so that this granularity becomes essential for our downstream methodology to learn relative stratification effectively, enabling correct reconstruction for 2.5D animation.
Formally, let $\mathcal{M} = \{m_1, m_2, \dots, m_M\}$ be the set of ArtMeshes in a model, and let $z(m_i)$ be the integer drawing order index of fragment $m_i$. We compute the normalised pseudo-depth $d(m_i) \in [0,1]$ by min--max normalisation, where $z_{\min}=\min_{m\in\mathcal{M}} z(m)$ and $z_{\max}=\max_{m\in\mathcal{M}} z(m)$, as $d(m_i)=(z(m_i)-z_{\min})/(z_{\max}-z_{\min})$.

To summarise, by applying this rigorous data pipeline, we successfully constructed a large-scale dataset comprising $9,102$ fully annotated 2.5D Live2D models after augmentation, curated from diverse community platforms including ArtStation, Booth, and DeviantArt. The dataset is partitioned into a training set of $7,404$ samples, a validation set of $851$, and a test set of $847$ specifically reserved for the evaluation of layer separation and depth inference. The total labelling time is around $12$ hours. We commit to releasing the full annotation codebase, the custom verification GUI, and the pre-trained 2D segmentation model to the research community.

\section{Methodology}
\label{sec:methodology}

We propose a framework that couples high-fidelity, globally consistent RGBA generation of distinct body parts with an explicit drawing-order signal, providing a unified pipeline for 2.5D-ready character reconstruction.
\begin{figure*}[!t]
    \centering
    \includegraphics[width=0.8\linewidth]{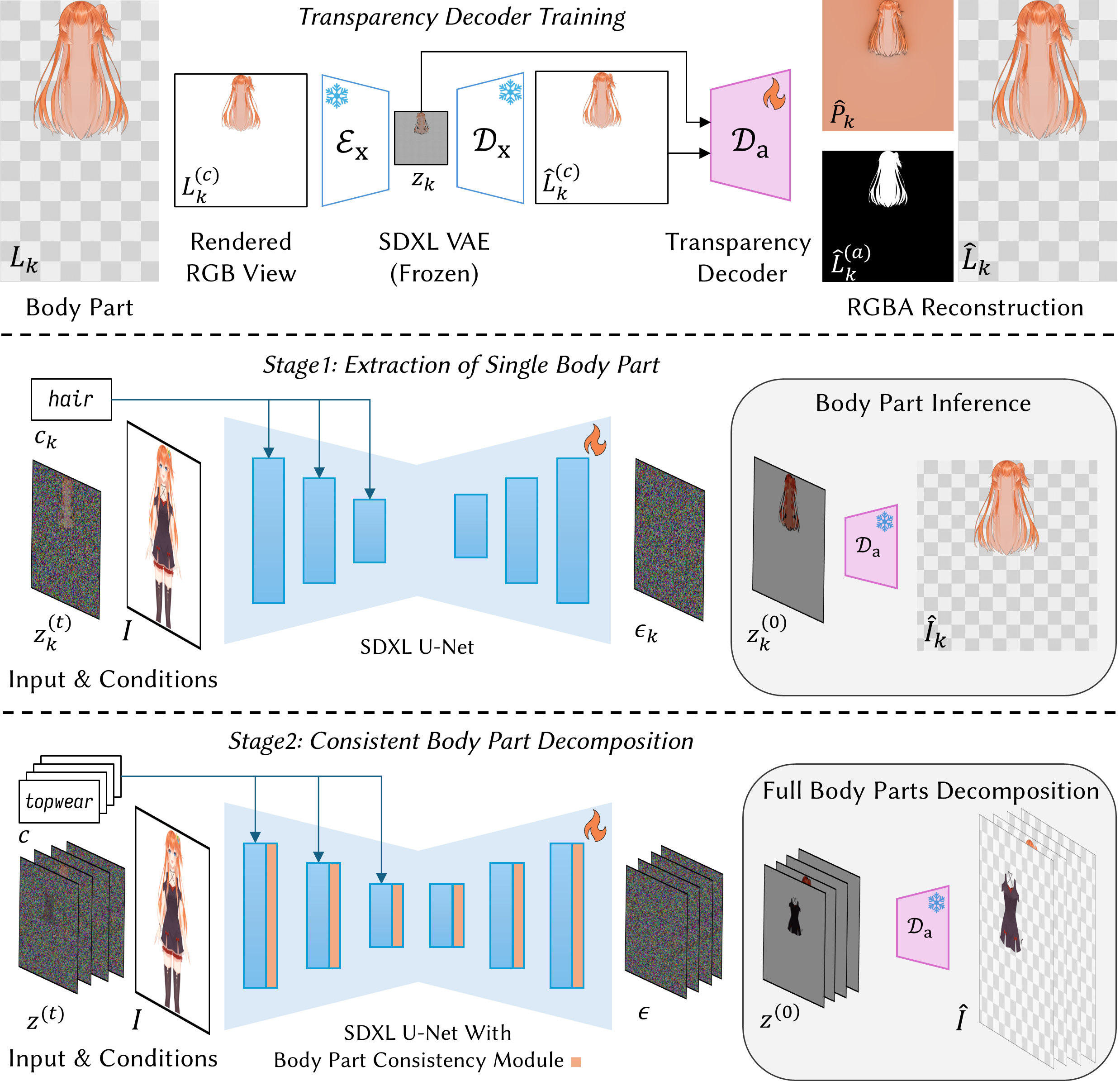}
    \caption{Training process for our body part decomposition framework.}
    \label{fig:methodology}
\end{figure*}

\subsection{Semantic Layer Decomposition}

We propose the semantic body part decomposition framework to decompose a single static anime illustration $I \in \mathbb{R}^{H \times W \times 3}$ into a comprehensive set of $N=19$ semantic layers $\mathcal{L}=\{L_1,\dots,L_N\}$. Each layer $L_k \in \mathbb{R}^{H \times W \times 4}$ is synthesised as a fully inpainted RGBA image, capturing both visible surfaces and hallucinated occluded regions necessary for animation.
To achieve the high-fidelity output required for this domain, we architect our solution upon the SDXL~\cite{sdxl} backbone. We strategically leverage this foundation model to harness its native $1024{\times}1024$ resolution and robust generalisation, ensuring the preservation of the intricate, high-frequency details characteristic of anime art.

\subsubsection{Transparency Decoder For Body Parts}
To enable transparent-layer generation within our framework, we adapt the Latent Transparency strategy from LayerDiffusion~\cite{layerdiffusion}.
For each semantic layer of body parts $L_k$, we denote its RGB channels as $L_k^{(c)}\in\mathbb{R}^{H\times W\times 3}$ and its alpha channel as $L_k^{(\alpha)}\in\mathbb{R}^{H\times W\times 1}$. Since RGB values are undefined where $\alpha=0$, we follow the iterative Gaussian blurring convention to fill transparent regions of $L_k^{(c)}$, to form a padded Gaussian $P_k$ which improves boundary anti-aliasing and layer blending during compositing.
Since we do not require transparent input for the diffusion process, we keep the SDXL VAE encoder $\mathcal{E}_{\text{x}}$ and decoder $\mathcal{D}_{\text{x}}$ untouched. Concretely, we encode the direct RGB component of the body part into a latent $z_k=\mathcal{E}_{\text{x}}(L_k^{(c)})$ and then reconstruct it with the SDXL decoder, as $\hat{L}_k^{(c)}=\mathcal{D}_{\text{x}}(z_k)$.
To enable RGBA-compatible decoding, we introduce a trainable transparency decoder $\mathcal{D}_{\text{a}}$, which is conditioned on both $z_k$ and $\hat{L}_k^{(c)}$ to predict the padded Gaussian image and the alpha $[\hat{P}_k,\hat{L}_k^{(\alpha)}]=\mathcal{D}_{\text{a}}(z_k,\hat{L}_k^{(c)})$. Together, these predicted components are combined to produce the final RGBA reconstruction of the body part $\hat{L}_k$. In this design, we preserve the original SDXL latent-space distribution, which simplifies subsequent diffusion fine-tuning, and we delegate transparency entirely to the transparency decoder.

We initialise the transparency decoder $\mathcal{D}_{\text{a}}$ from the \textit{sd-forge-layerdiffuse} \cite{sd-forge-layerdiffuse} repository, which provides a strong prior for transparency in general photography and graphical designs. We train on our dataset using an RGBA reconstruction loss with an adversarial term:
\begin{equation}
    \mathcal{L}_{\text{dec}} =
    \left\lVert L_k^{(c)} - \hat{L}_k^{(c)} \right\rVert_2
    + \left\lVert L_k^{(\alpha)} - \hat{L}_k^{(\alpha)} \right\rVert_2
    + \lambda_{\text{disc}}\mathcal{L}_{\text{disc}},
    \label{eq:decoder}
\end{equation}
where $\lVert\cdot\rVert_2$ denotes the $L_2$ norm, $\mathcal{L}_{\text{disc}}$ is a PatchGAN discriminator loss for sharper boundaries and improved perceptual quality, and we set $\lambda_{\text{disc}}=0.01$.

\begin{figure*}[!t]
    \centering
    \includegraphics[width=0.9\linewidth]{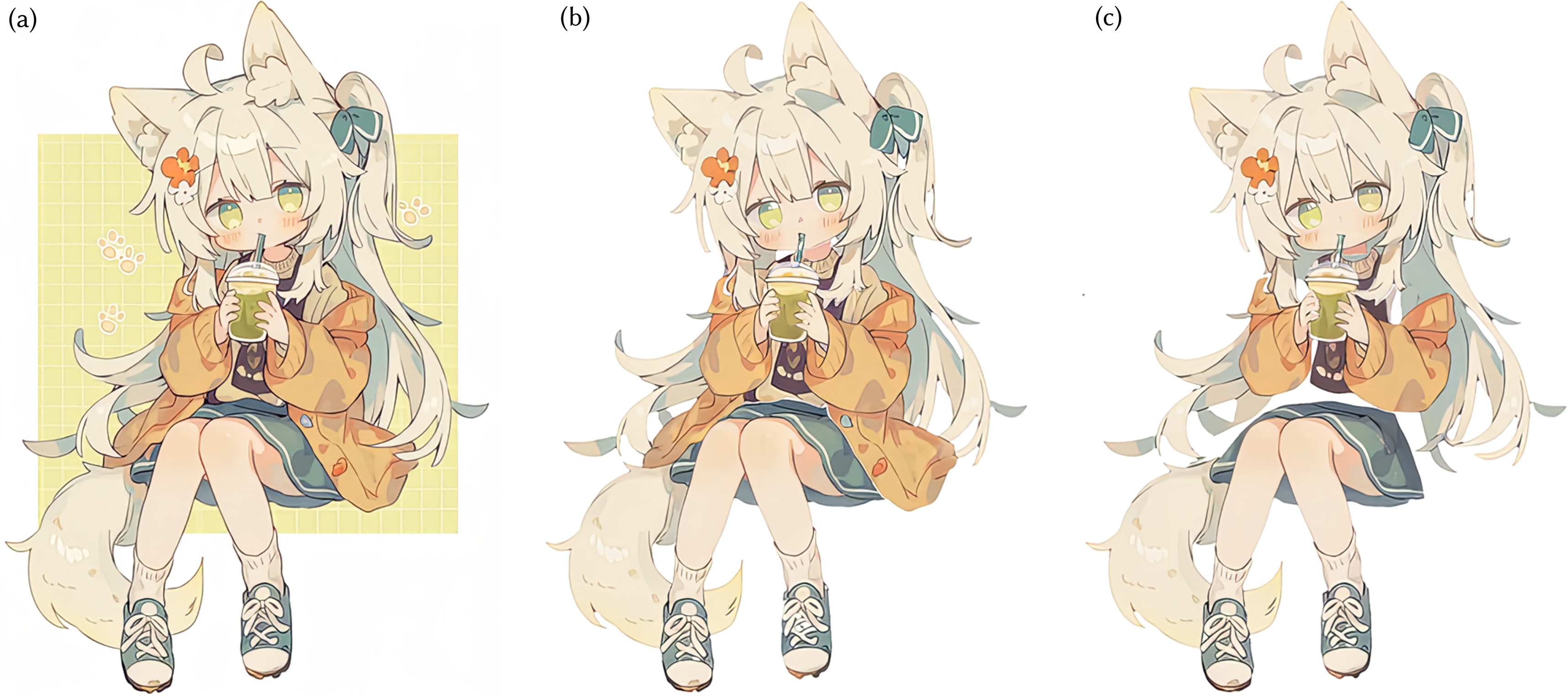}
    \caption{Visualization of our Body Part Consistency Module. (a) Input; (b) Reconstruction with the Module; (c) Reconstruction without the Module. Without the module, the body part decomposition tend to be incomplete.}
    \label{fig:attention_visualization}
\end{figure*}

\subsubsection{Diffusion Model for Body Parts Decomposition}

With transparent layer decoding enabled, we next fine-tune the diffusion backbone to generate semantic body-part layers. We adopt a two-stage local-to-global training strategy.

\paragraph{The first stage.} In this stage, we primarily reduce the domain gap of vanilla SDXL and teach the model to extract \emph{one} target part at a time with high fidelity.
Given an input illustration $I \in \mathbb{R}^{H \times W \times 3}$ and a target class $c_k$, we fine-tune the SDXL U-Net as a conditional denoiser to extract the class-specified RGBA body part. Instead of using free-form text prompts, we represent each semantic class $c_k$ with a discrete embedding using an encoder $\tau_C(c_k)$ and inject it via cross-attention~\cite{stablediffusion}. We also condition on the source illustration through an image-conditioning encoder as $\tau_I(I)$ via zero-convolution into the encoder layers~\cite{controlnet}. For each target layer $L_k$, we encode its RGB channels as $z_k^{(0)}=\mathcal{E}_{\text{x}}\!\left(L_k^{(c)}\right)$, sample a diffusion timestep $t$ and $\epsilon\sim\mathcal{N}(0,\mathbf{I})$, and form the noisy latent $z_k^{(t)}=\alpha_t z_k^{(0)}+\sigma_t \epsilon$ using the standard noise schedule. The denoising U-Net is trained with $\epsilon$-prediction:
\begin{equation}
\mathcal{L}_{\text{stage1}}
=
\mathbb{E}_{I,\, k,\, t,\, \epsilon}
\left[
\left\|
\epsilon -
\epsilon_\theta\!\left(
z_k^{(t)},\, t;\, \tau_I(I),\, \tau_C(c_k)
\right)
\right\|_2^2
\right],
\label{eq:stage1}
\end{equation}
where layer index $k$ is sampled uniformly from $\{1,\dots,N\}$, $t$ is sampled uniformly from diffusion timesteps, and $\epsilon$ is standard Gaussian noise. This stage produces locally plausible, high-fidelity part extractions in the anime domain.

\paragraph{The second stage.} While the first stage produces high-quality predictions when extracting one body part at a time, it often fails to decompose \emph{all} parts reliably. In particular, the model may leave some semantic layers nearly empty, effectively assigning ambiguous regions to other, visually adjacent parts. An example is demonstrated in Figure~\ref{fig:attention_visualization}, where the upper-body clothing has a similar colour and texture to the sleeves, causing it to be ignored.

We hypothesise that, without a global view of the status of decomposition, the model may fail to explicitly reason about which regions have already been explained by other decomposed layers, and thus fails to allocate content consistently across all $N$ body parts. To enforce global consistency over the entire layer stack, we reformulate the decomposition of body parts as a \emph{joint} denoising problem.
Concretely, instead of predicting each $z^{(k)}$ independently, we stack all $N$ part latents along a new part dimension $c$, analogous to the temporal dimension in video diffusion~\cite{svd, videocrafter}, and denoise them simultaneously. We denote the stacked latent tensor at diffusion timestep $t$ as $Z^{(t)}=[z_1^{(t)},\dots,z_N^{(t)}]$, where the $k$-th slice corresponds to semantic layer $L_k$. We retain class conditioning in this stage: each slice along $c$ corresponds to a semantic layer, and we inject its class embedding $\tau_C(c_k)$ via cross-attention following the first-stage convention, while keeping the same image conditioning $\tau_I(I)$, where we denote the set of class labels as $C=\{c_1,\dots,c_N\}$.

To enable information exchange across parts during denoising, we insert a \textbf{Body Part Consistency Module} after each spatial attention block in the U-Net. This module performs attention along the part dimension $c$, allowing each layer to condition on the current predictions of all other layers. Since semantic layers are largely independent and only couple through occlusion boundaries, we do not introduce any convolution along $c$. Instead, the module provides a \emph{global} mechanism that helps distribute ambiguous content across layers, while the per-slice class embedding $\tau_C(c_k)$ acts as a \emph{local} semantic constraint that keeps each slice aligned with its designated body part.
For training, we initialise from the first-stage U-Net and jointly optimising both the original U-Net parameters and the newly inserted Body Part Consistency Module. The training objective remains the standard $\epsilon$-prediction loss, but applied to the stacked latent tensor $Z^{(t)}$:
\begin{equation}
\mathcal{L}_{\text{stage2}}
=
\mathbb{E}_{I,\, t,\, \epsilon}
\left[
\left\|
\epsilon -
\epsilon_\phi\!\left(
Z^{(t)},\, t;\, \tau_I(I),\, \tau_C(C)
\right)
\right\|_2^2
\right].
\label{eq:stage2}
\end{equation}
After training, we jointly sample the full latent stack and decode each part with the frozen SDXL decoder $\mathcal{D}_{\text{x}}$, then apply our transparency decoder $\mathcal{D}_{\text{a}}$ to obtain the final set of decomposed RGBA layers.

\begin{figure*}
    \centering
    \includegraphics[width=\linewidth]{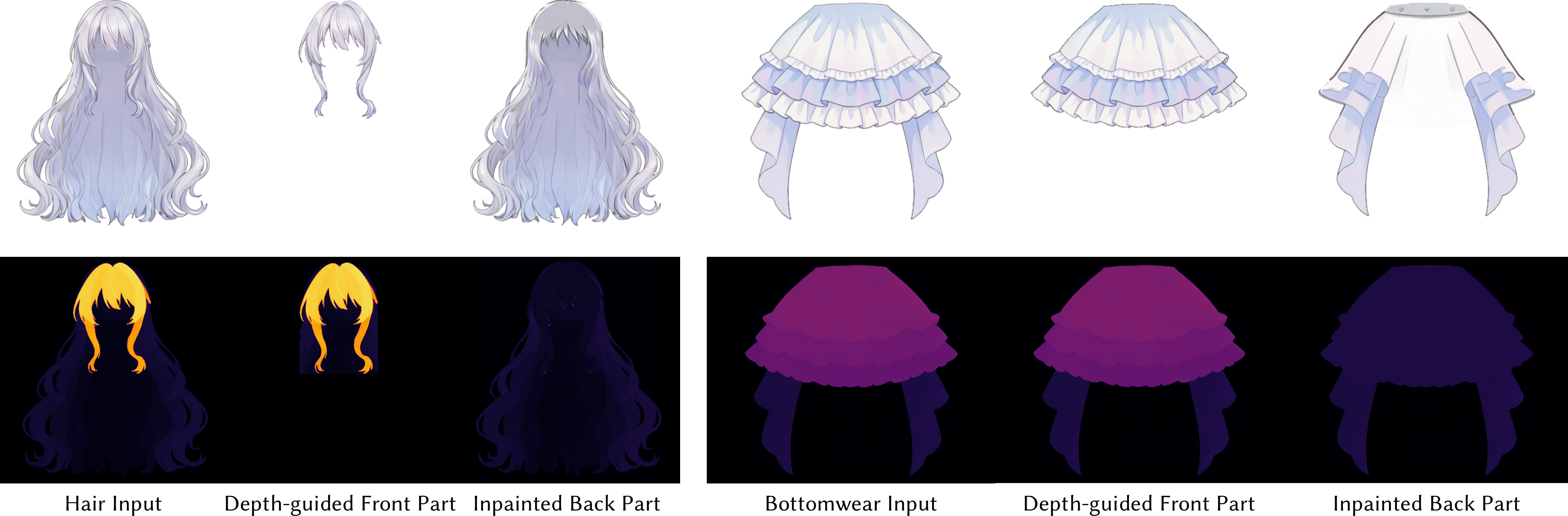}
    \caption{Depth-guided stratification within a semantic layer. We cluster pseudo-depth values into front/back strata and then inpaint the newly exposed regions.}
    \label{fig:stratification}
\end{figure*}

\subsection{Reconstruction of the Decomposed Layers}

\subsubsection{Drawing-Order Inference using Pseudo-Depth}
To infer scene-consistent drawing order under complex, interleaving occlusions, we predict a dense pixel-level pseudo-depth map $D_k$ for each semantic layer. We build on Marigold~\cite{marigold}, a state-of-the-art affine-invariant depth estimator, as the backbone. Marigold provides strong geometric priors for estimating relative depth, which we find also effective for drawing-order inference in anime compositing, even though our pseudo-depth does not represent metric 3D depth.

We adopt a two-stage approach similar to our body part decomposition pipeline, except that we do not require the special transparent VAE in this step.
In the first stage, we fine-tune Marigold to predict the pseudo-depth map $D_k$ for a specific semantic region conditioned on the tag $c_k$. This allows the model to adapt its geometric priors to the specific topology of anime parts, such as the curvature of hair strands relative to the face.
After that, in the second stage, we address the issue of inconsistent scale across independently predicted maps. We stack the latents of all 19 depth maps and re-employ the Body Part Consistency Module to enforce attention across the part dimension. This establishes a unified, globally ordered pseudo-depth hierarchy, ensuring that the relative depths of interacting parts are logically consistent and ready for reassembly.

\subsubsection{Depth-guided Layer Stratification}
Given the predicted RGBA layers and their pseudo-depth, we can in principle composite the image by z-buffering. However, this strategy fragments a semantic part into many disconnected pieces, which is incompatible with 2.5D pipelines that require coherent, manipulatable layered slices. A simpler alternative is to treat each semantic part as a whole by assigning it a representative depth value and then z-sorting the parts. Yet, this part-wise ordering cannot express intra-class interleaving, leading to the ``sandwich'' case where a single semantic layer must appear both in front of and behind another layer (e.g., hair vs. face).

To retain layer-based outputs while supporting such intra-class stratification, we shall subdivide a semantic layer into a small number of depth strata. As revealed in Figure~\ref{fig:stratification}, we observe that the predicted pseudo-depth often forms separated modes within self-occluding parts (e.g., fore-hair vs.\ back-hair). In this way, we apply K-Means clustering to pseudo-depth values inside the alpha mask of selected classes for further subdivision. By default, we apply a $K{=}2$ K-Means clustering to split these classes into front/back strata.

In our applications that export Photoshop PSD layers, we apply this heuristic to \textit{Hair}, \textit{Handwear}, \textit{Topwear}, and \textit{Bottomwear}. Because subdivision introduces new occlusion boundaries, we additionally run a generic inpainting step to fill newly uncovered regions; in practice, we find a modern inpainter such as LaMa~\cite{lama} works well. Note that this inpainting can contain minor mistakes, but they typically lie in the far-back stratum of the ``sandwich'' and are difficult to observe after compositing.

\section{Experiments}
\label{sec:experiments}

We evaluate our framework against relevant baselines and demonstrate its effectiveness for downstream animation applications. Implementation details and training hyperparameters are provided in the supplementary material.

\subsection{System Evaluation}

\begin{figure*}
    \centering
    \includegraphics[width=\linewidth]{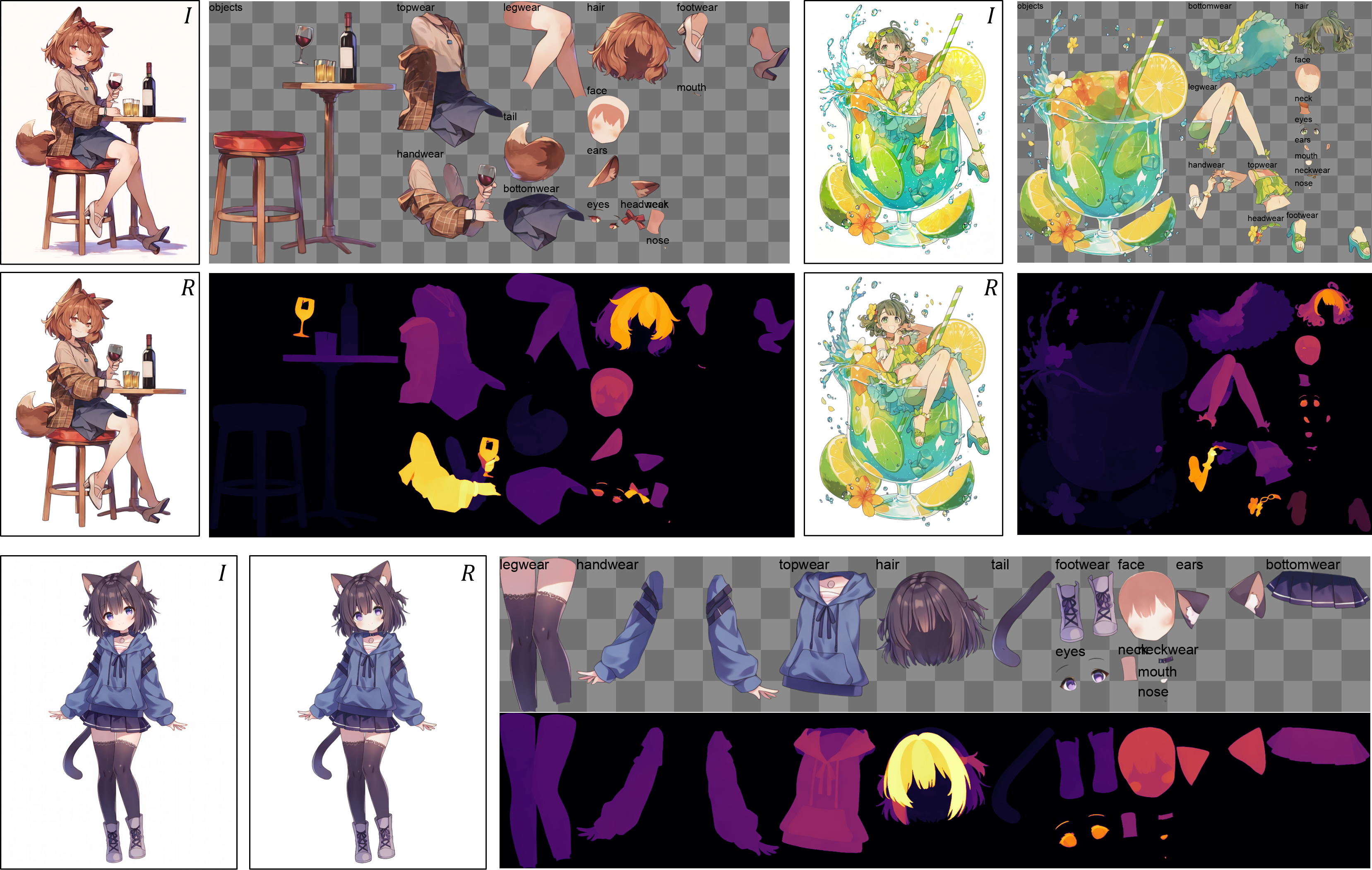}
    \caption{Showcase results. For each example, we show the input illustration ($I$), our decomposed semantic RGBA layers, the predicted pseudo-depth, and the reconstructed composite ($R$). The top-left example contains a minor artefact where the wine glass is duplicated, which can be easily corrected in a layer editor.}
    \label{fig:showcase}
\end{figure*}

\begin{figure*}
    \centering
    \includegraphics[width=\linewidth]{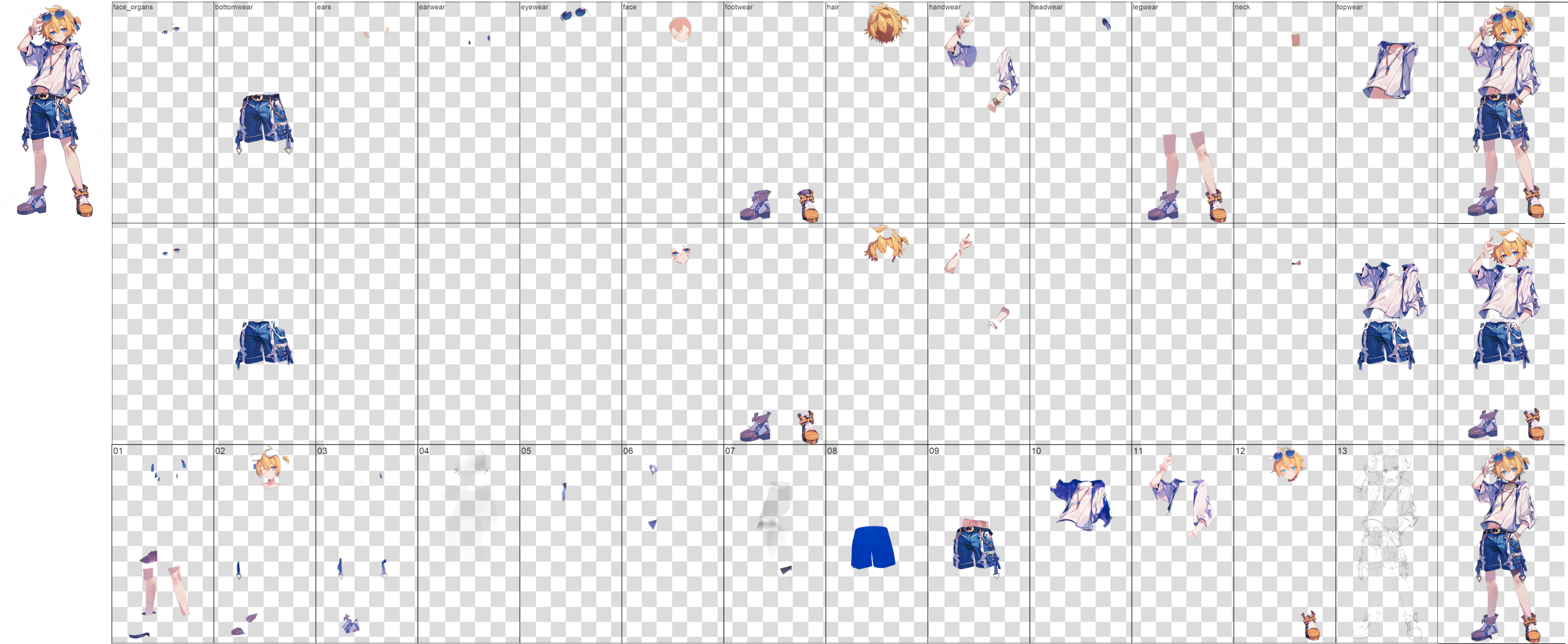}
    \caption{Visual comparison with layer decomposition baselines. (a) Ours (2.5D); (b) SAM3 (2D); (c) Qwen-Image-Layered (2.5D). Ours and SAM3 predict part semantics (we merge facial parts for visualisation), whereas Qwen-Image-Layered uses a single text prompt. The rightmost column shows the reconstruction.}
    \label{fig:visual_comparison}
\end{figure*}

\begin{table*}[!t]
\centering
\caption{Quantitative evaluation of our framework against baselines. Arrows indicate whether lower ($\downarrow$) or higher ($\uparrow$) values are better.}
\label{tab:quantitative_results}
\resizebox{0.7\linewidth}{!}{%
\begin{tabular}{l|ccc|ccc}
\toprule
Method & LPIPS $\downarrow$ & PSNR $\uparrow$ & SSIM $\uparrow$ & \makecell{Mask \\ Dice loss $\downarrow$} & \makecell{Mask \\ MSE $\downarrow$} & FID $\downarrow$ \\
\midrule
\makecell[l]{SAM+LaMa} & 0.2880 & 12.2802 & 0.8445 & 0.4336 & 0.1020 & 81.1419 \\
\midrule
\makecell[l]{Ours without\\Consistency Module} & 0.1952 & 16.2350 & 0.9053 & 0.6480 & 0.0640 & \textbf{16.7069} \\
\midrule
Ours Full & \textbf{0.1549} & \textbf{18.2965} & \textbf{0.9230} & \textbf{0.3855} & \textbf{0.0354} & 18.3700 \\
\bottomrule
\end{tabular}%
}
\end{table*}

\subsubsection{Layer Decomposition and Reconstruction}
By ``seeing through'' a single 2D illustration, our proposed framework achieves plausible completion of occluded regions and fine-grained stratified ordering cues. We demonstrate representative results in Figure~\ref{fig:showcase}, where the predicted layers and pseudo-depth enable near-perfect reconstruction of the input character, with hallucinated hidden parts ready for large deformations in animation. We omit background reconstruction, since production pipelines typically replace it with a new scene. We strongly recommend that readers refer to the supplementary material and video for additional qualitative results.

Additionally, we compare our approach against the state-of-the-art Qwen-Image-Layered~\cite{qwen-image-layered}. We exclude LayerDiffusion since its original design makes it difficult to scale training to a similar number of layers as ours. Qwen-Image-Layered also requires text prompts to specify layer semantics\footnote{We use the prompt \textit{separate into different body part layers (include eyes, hair, arms, ears, torso, legs, etc), which will be reconstructed to a full body illustration.} to guide it towards our decomposition setting.} As illustrated in Figure~\ref{fig:visual_comparison}, Qwen exhibits significant limitations: it struggles to precisely extract specific body parts, often merging distinct semantic layers or inconsistently fragmenting a single part into separate RGBA regions. While its reconstruction loss is low due to hard constraints, the intermediate layers are incoherent and unusable for animation. We attribute this to its focus on top-to-bottom graphic layer generation, which does not model the stratified, interleaving occlusions common in anime characters. Consequently, we omit a quantitative comparison for this baseline.
We also considered object-level scene deocclusion methods such as~\cite{object-level-scene-deocclusion}. However, they are not designed to model the fine-grained stratification required by anime character layering, as discussed in Section~\ref{sec:related_work}, which makes a direct evaluation less meaningful\footnote{\cite{object-level-scene-deocclusion} did not release code, and we were unable to reproduce their results. We find retraining their VAE on anime data caused a large distribution shift, which destabilised the pretrained diffusion backbone and led to training collapse.}.
Regarding inference latency, the layer decomposition requires approximately $74$ seconds per input to generate $1024\times1024$ layered images, while depth inference takes $10$ seconds on a single NVIDIA RTX 4090 GPU. Given that this is a one-time preprocessing step for any given asset, the computational cost is within acceptable limits for production workflows.

\begin{figure*}
    \centering
    \includegraphics[width=\linewidth]{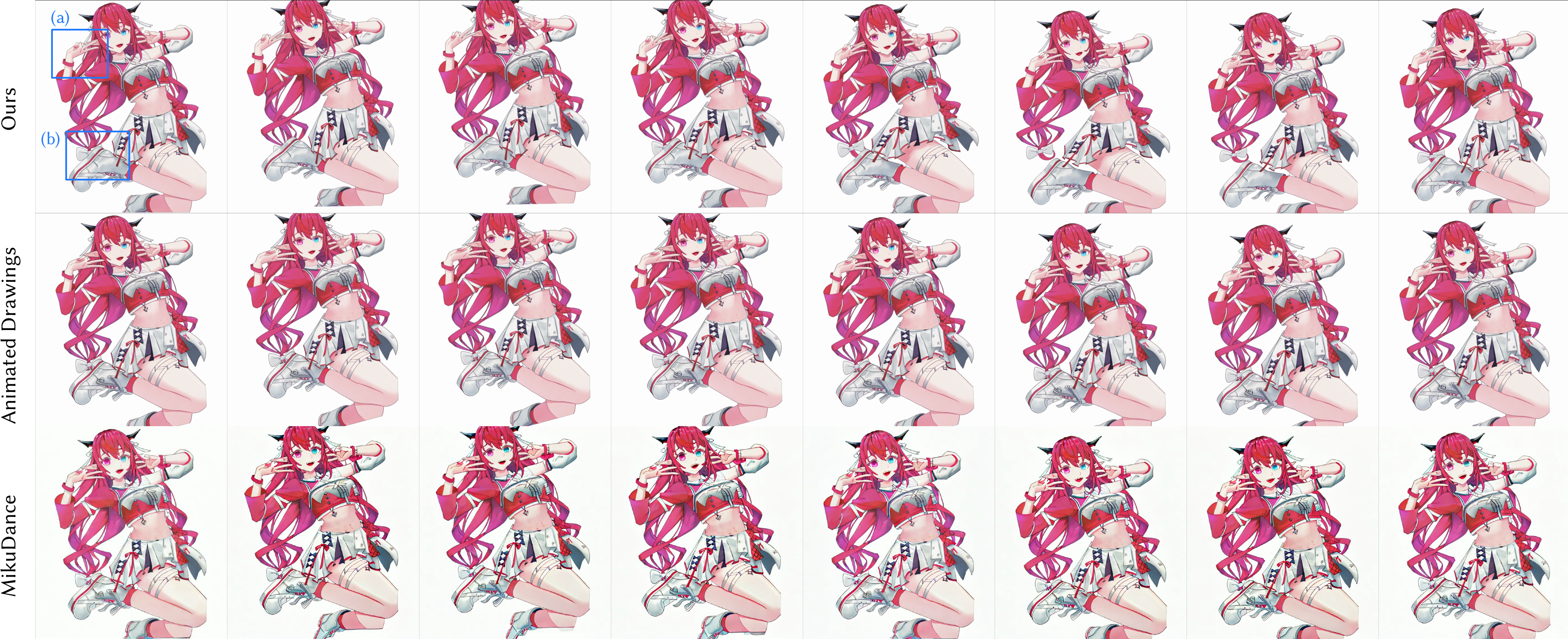}
    \caption{Fully automatic puppet animation driven by a video sequence. Insets highlight that our animation is free from (a) severe geometric distortion (e.g., tearing/stretching) and (b) preserves a stronger sense of depth and occlusion in complex regions. Refer to the supplementary video for clearer visualisation.}
    \label{fig:puppet_animation}
\end{figure*}

\begin{figure*}
    \centering
    \includegraphics[width=\linewidth]{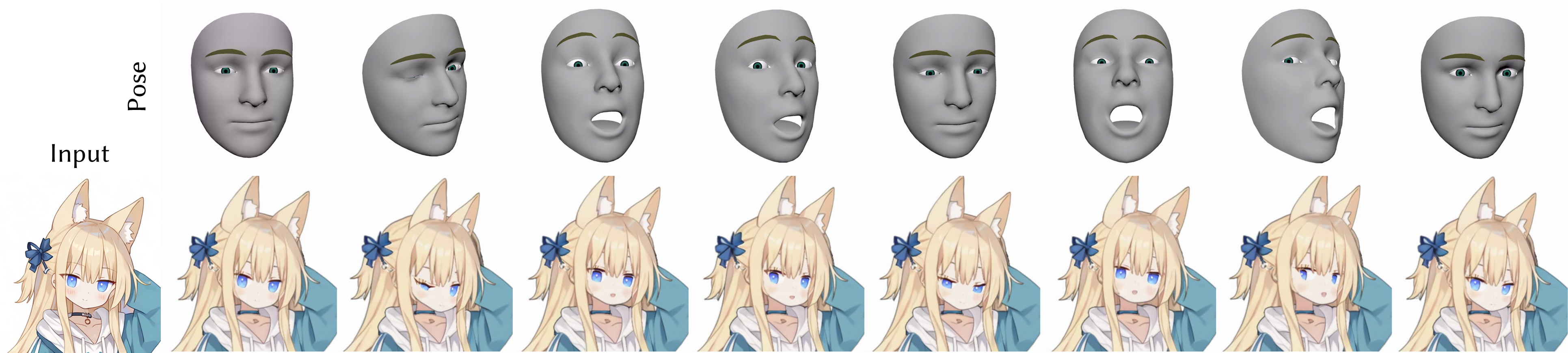}
    \caption{Real-time ``talking-head'' synthesis. We drive our decomposed layers using 3D facial landmarks (top). The resulting animation (bottom) demonstrates that our method successfully handles the changing occlusions required for head turning and facial deformation.}
    \label{fig:talking_head}
\end{figure*}

\subsubsection{2D Semantic Body Parsing}

Our framework naturally serves as a strong 2D semantic body parser, since it predicts part-level segmentation masks as an intermediate output. We therefore compare against SAM3~\cite{sam3}, a state-of-the-art prompted segmentation model. To enable a part-by-part comparison, we design prompts that match our 19-part taxonomy and report qualitative results in Figure~\ref{fig:visual_comparison}.
As illustrated, SAM3 often produces incomplete or ambiguous segmentations on anime characters. Typical failure cases include missing regions (e.g., hair or trousers) and overlapping masks across parts, since SAM3 does not enforce a globally consistent allocation of pixels to individual body parts. More fundamentally, SAM3 is a strictly planar 2D method and has no ``see-through'' capability to infer occluded body structure, which is essential for 2.5D-ready decomposition. In addition to qualitative comparisons, we perform a quantitative evaluation by measuring part-level segmentation accuracy including MSE and Dice loss~\cite{diceloss} against the mask annotations in our 2.5D test set and report the metrics in Table~\ref{tab:quantitative_results}.
Finally, we test whether a purely 2D pipeline can approximate our occlusion completion by combining segmentation with a modern inpainter. Concretely, we extract the ground-truth inverse visibility mask of body parts from our 2.5D test data, inpaint the missing regions with LaMa~\cite{lama} and then compare the results using reconstruction metrics and FID~\cite{fid}. We find that this baseline fails to recover plausible hidden anatomy since it lacks inter-part consistency cues and exhibits a large domain gap on anime textures and line work.

\subsubsection{Ablation Study of the Body Part Consistency Module}

We ablate the Body Part Consistency Module to quantify its effect on both (i) semantic layer decomposition and (ii) pseudo-depth prediction for drawing-order inference. As summarised in Table~\ref{tab:quantitative_results}, adding this module yields better scores on reconstruction metrics, indicating improved completeness and cross-layer coherence of the predicted parts. The slight degradation in FID is also expected, as enforcing stronger global consistency can reduce output diversity. Qualitative evidence is also provided in Figure~\ref{fig:attention_visualization}.
Beyond decomposition, this module also improves our pseudo-depth estimation used for ordering: on our test set, it reduces mean absolute relative error (AbsRel) from $0.1549$ to $0.0943$ and increases $\delta_1$ score from $0.8427$ to $0.9103$, leading to more globally consistent drawing-order inference.

\subsection{Applications}

\paragraph{Puppet Animations.}
Our framework enables stable 2.5D animation from a single illustration using deterministic 2D deformation and compositing, together with real-time physics (e.g., spring-based hair dynamics). Because the character is represented as stratified RGBA parts, individual components can be animated with independent motion and physics, producing a stronger sense of depth and parallax, as well as plausible re-appearance of previously occluded regions during articulation. We integrate our stratified layers into an Animated Drawings~\cite{animateddrawings} pipeline and compare against its vanilla implementation in Figure~\ref{fig:puppet_animation}. Under the same driving motion, our method maintains coherent occlusions and layer relationships during large deformations: for example, it preserves plausible skirt--shoe--hair stratification (Figure~\ref{fig:puppet_animation}(b)) and avoids the tearing/stretching artifacts visible in Animated Drawings, which become pronounced in the 3rd and 4th columns (Figure~\ref{fig:puppet_animation}(b)). We also compare with the end-to-end generation method MikuDance~\cite{mikudance}; while it can synthesize motion, it still exhibits temporal inconsistency and local distortions, and it does not provide real-time performance. 
To assess practical usability, we also export our results as layered PSD files and asked professional animators to create keyframes; they reported that the outputs are largely production-ready and require only minor manual modifications. We show these results in the supplementary material.

\paragraph{Talking-head VTubing.}
We also implement a real-time ``talking-head'' system for VTubing based on our stratified facial and hair layers and we demonstrate results in Figure~\ref{fig:talking_head}. The separated RGBA parts enable stable control of expressions and head motions through 2D deformers, while preserving fine anime details (line work and flat shading). The stratified ordering and occlusion completion reduce ambiguities when applying large facial deformations, enabling accurate pose following while preserving 2D aesthetic realism.

\section{Conclusion}

In this work, we present the framework that converts a single anime illustration into a 2.5D-ready character model by decomposing the input into complete semantic RGBA body part layers and predicting a stratified drawing order. With a solid bootstrapped dataset and a cross-part decomposition model, our framework produces aesthetically faithful decompositions and enables near-perfect reconstruction, supporting practical downstream animation workflows.

Our framework is not without limitations. We occasionally observe minor overlaps between predicted layers when they lie outside the body (as in Figure~\ref{fig:showcase}, top left), but these local artefacts are easy to fix in standard layer editors. Additionally, depth-guided stratification is difficult to split into multiple stable sublayers, since the pseudo-depth may not always induce reliable discrete boundaries. We plan to address these issues in future work, together with predicting automatic animation timings from our decomposition.

{
    \small
    \bibliographystyle{ieeenat_fullname}
    \bibliography{sample-base}
}


\clearpage
\appendix

\section{Additional Information for the 2.5D-level Annotation and Refinement}

When the 2D annotation is bootstrapped, we use the fragment visibility maps of the ArtMeshes to project the predicted semantic labels back onto the individual texture fragments. Specifically, for any visible fragment, we assign the initial semantic class based on the majority vote of the pixel-wise predictions falling within its visible region. However, due to the inherent nature of Live2D compositing, many fragments may be fully occluded in the reference pose (e.g., the back of the hair, or limbs hidden behind clothing) and thus receive no direct supervision from the 2D segmentation result.

To resolve this, we propose a hierarchical propagation strategy to infer the semantic labels of hidden fragments. We apply a three-stage heuristic:
\begin{itemize}
    \item \textbf{Semantic String Matching:} We first exploit the original artist's naming conventions. If an unlabelled ArtMesh contains a specific keyword (e.g., ``PartHair'', ``PartHand''), which is also a common substring with already labelled fragments, we propagate the known annotation to the unlabelled mesh.
    \item \textbf{Sibling Voting:} Live2D model ArtMeshes are hierarchically structured. If string matching is inconclusive, we perform a majority vote based on the ArtMeshes residing at the same hierarchical level (siblings), assuming that fragments grouped together likely share the same semantic class.
    \item \textbf{Recursive Parent Voting:} If the immediate hierarchical level lacks labelled data, we recursively query the parent levels until a meaningful source for majority voting is found.
\end{itemize}

\begin{figure*}[!t]
    \centering
    \includegraphics[width=\linewidth]{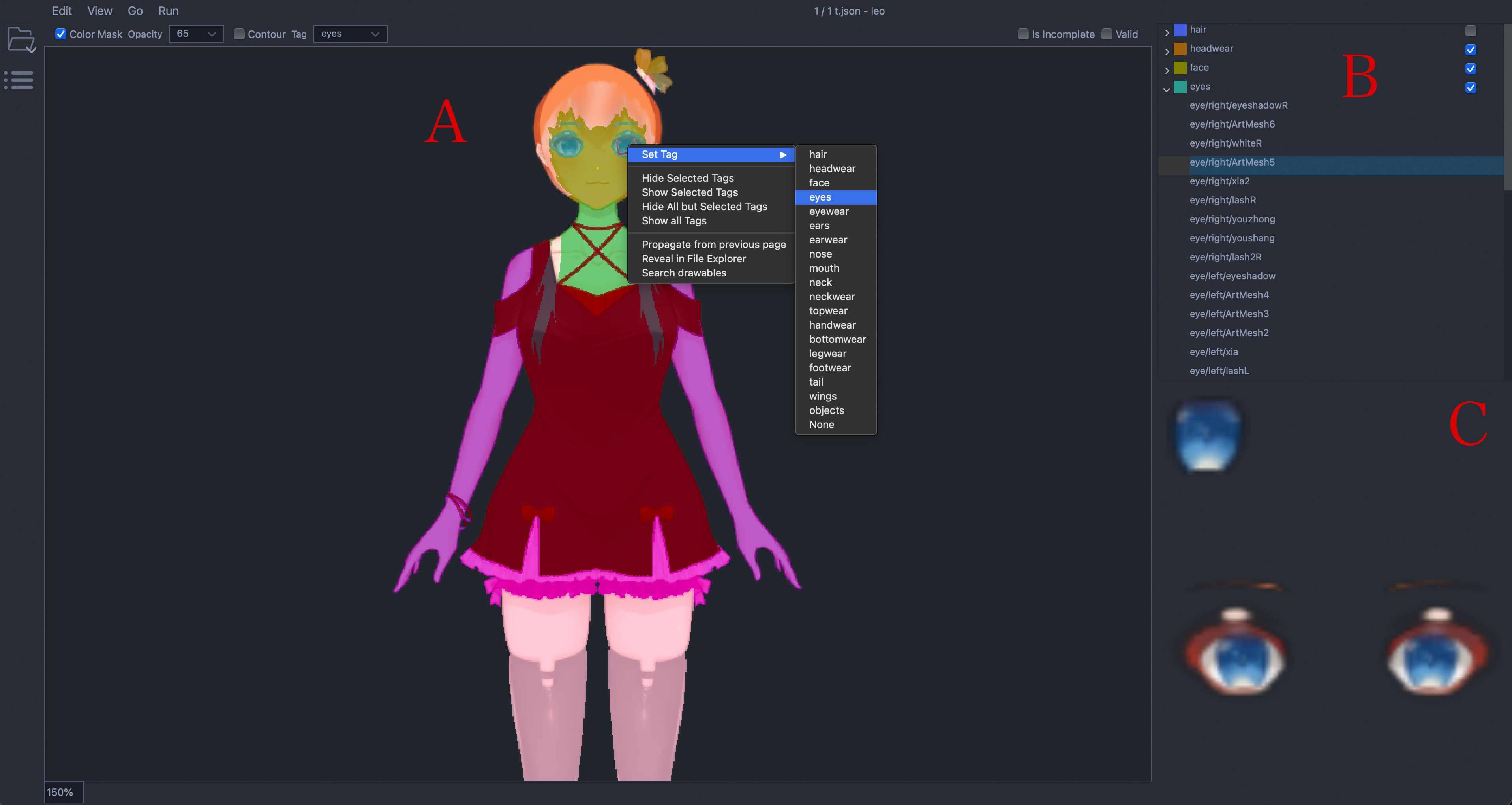}
    \caption{Screenshot of our GUI for Live2D model labelling. (A) Main preview panel; (B) ArtMesh list grouped by assigned class; (C) Preview of the selected ArtMesh and its semantic group (in this case, eyes). We disable hair for clearer visualisation.}
    \label{fig:dataset_gui}
\end{figure*}

To ensure ground-truth quality, we developed a custom graphical user interface for manual verification and refinement of Live2D model labelling, as shown in Figure~\ref{fig:dataset_gui}. The interface allows annotators to inspect labels with ``see-through'' capabilities, offering tools to zoom and toggle visibility of arbitrary ArtMesh fragments (Figure~\ref{fig:dataset_gui}(A)).
For usability, we allow users to toggle the visibility of ArtMesh groups based on their \textbf{body part class} or \textbf{ArtMesh hierarchy} (Figure~\ref{fig:dataset_gui}(B)). Together with the preview of the selected ArtMesh and all ArtMeshes with the same semantic labelling (Figure~\ref{fig:dataset_gui}(C)), the GUI allows rapid checking of body part completeness and identification of false positives (e.g., when visualising \textit{hand} fragments only, any fragments outside the hand region must be false positives). This workflow enables efficient correction of deep, occluded layers with much less effort.

\section{Additional Baselines and Statistics}

\paragraph{Mask-conditioned baseline (SAM mask as condition).}
Inspired by the mask-prompted completion strategy in~\cite{object-level-scene-deocclusion}, we additionally test whether a spatial prompt alone can improve single-layer extraction. Concretely, in Stage~1 we discard the class-based conditioning (i.e., our semantic class embedding) and instead condition the model on the visible-region mask predicted by our fine-tuned 2D SAM model. Intuitively, this mask provides a stronger spatial cue than text. However, as shown in Table~\ref{tab:quantitative_results_supp}, this variant performs substantially worse and produces less coherent intermediate layers.
We believe this failure stems from the lack of explicit semantics in the prompt: in anime characters, multiple parts can be spatially adjacent or overlap in projection, and a mask alone does not uniquely specify which semantic layer the model should extract, leading to ambiguous and inconsistent allocation.

\begin{table*}[!t]
\centering
\small
\setlength{\tabcolsep}{6pt}
\caption{Additional quantitative evaluation against baselines. Arrows indicate whether lower ($\downarrow$) or higher ($\uparrow$) values are better.}
\label{tab:quantitative_results_supp}
\begin{tabular}{l|ccc|ccc}
\toprule
Method & LPIPS $\downarrow$ & PSNR $\uparrow$ & SSIM $\uparrow$ & \makecell{Mask \\ Dice loss $\downarrow$} & \makecell{Mask \\ MSE $\downarrow$} & FID $\downarrow$ \\
\midrule
\makecell[l]{Ours without\\Consistency Module} & 0.1952 & 16.2350 & 0.9053 & 0.6480 & 0.0640 & \textbf{16.7069} \\
\midrule
Ours Full & \textbf{0.1549} & \textbf{18.2965} & \textbf{0.9230} & \textbf{0.3855} & \textbf{0.0354} & 18.3700 \\
\midrule
\makecell[l]{Mask-conditioned\\(SAM mask)} & 0.2143 & 14.6300 & 0.9090 & 0.8469 & 0.1470 & 74.1500 \\
\midrule
\makecell[l]{SAM+LaMa} & 0.2880 & 12.2802 & 0.8445 & 0.4336 & 0.1020 & 81.1419 \\
\bottomrule
\end{tabular}
\end{table*}

\section{Training Details}
\label{sec:training_details}

We summarise the main training settings for each component of our framework below. Unless otherwise stated, we use AdamW optimisation with standard mixed-precision training.

\subsection{Multi-Decoder SAM Fine-tuning}
We fine-tune the multi-decoder SAM model on $4\times$ NVIDIA RTX 4090 GPUs for approximately $68$ hours. We train for $16{,}000$ steps with a batch size of $32$. We use AdamW with learning rate $2\times10^{-4}$ and weight decay $0.1$.

\subsection{Body Part Layer Diffusion Training}
We train our diffusion model for semantic RGBA body part generation at a resolution of $1024\times1024$.

\textbf{Stage~1 (local part extraction).}
We fine-tune the model on $8\times$ NVIDIA H200 GPUs for approximately $24$ hours. We train for $20{,}000$ steps with batch size $64$, using AdamW with learning rate $4\times10^{-5}$ and weight decay $0.01$, and a linear warm-up of $500$ steps.

\textbf{Stage~2 (joint denoising with global consistency).}
We train the full model (including the Body Part Consistency Module) on $8\times$ NVIDIA H200 GPUs for approximately $129$ hours. We use the same optimiser and hyperparameters as Stage~1, and train for $20{,}000$ steps at $1024\times1024$ resolution with batch size $64$.

\subsection{Pseudo-depth Training}
We fine-tune the pseudo-depth predictor (Marigold~\cite{marigold}) at a resolution of $768\times768$, using the same optimiser and hyperparameters as the diffusion training above (AdamW, learning rate $4\times10^{-5}$, weight decay $0.01$, warm-up $500$ steps, batch size $64$).

\textbf{Stage~1 (per-part pseudo-depth).}
We train on $8\times$ NVIDIA H200 GPUs for approximately $16$ hours.

\textbf{Stage~2 (joint pseudo-depth with global consistency).}
We train on $8\times$ NVIDIA H200 GPUs for approximately $115$ hours.

\section{Feedback from Artists}

We conducted an informal expert review with seven anime artists to assess the practical usability of our decomposed layers in real production settings. We provided each artist with the exported PSD files produced by our framework and asked them to (i) review the semantic correctness and editability of the extracted layers (including occlusion completion), and (ii) author animation timings and representative visual effects based on the layer stack, following a standard 2.5D workflow. Six artists returned completed feedback and animation drafts. Overall, their responses were positive: once we clarified that the PSDs were produced by an AI model, multiple artists reported that they did not expect an automatic system to produce a layered representation that is sufficiently structured for direct animation work, especially in terms of ``see-through'' completion of hidden regions in a fully automatic manner. Several artists described the decomposition as a useful starting point that can either be used directly for prototyping, or serve as a strong base that reduces manual preparation time when aiming for higher-quality, artist-refined layer stacks. Based on their submissions, artists were typically able to produce a high-quality motion draft within approximately 30--60 minutes, and we include representative results in the supplementary video. We also include a manga-style example, where the motion design suggests that the framework can be applied to sketch-like or more abstract illustration styles.

The artists also highlighted concrete limitations that align with our quantitative and qualitative observations. They reported that large, coherent structures (e.g., torso regions, hats, and major garments) tend to work well as independent layers, whereas thin and high-frequency details (e.g., sharp hair tips, small decorations, and fine accessories) sometimes require cross-checking the original illustration to resolve local ambiguities. In these cases, the layered output remains helpful, but does not fully eliminate manual judgement. In addition, some artists noted subtle ``AI-like'' visual patterns in a small subset of outputs, which can affect perceived line quality or texture regularity under close inspection. They also requested more fine-grained separations for certain semantics, for instance, explicitly separating left/right arms, to further reduce manual editing in Photoshop or similar tools, even if a coarser separation is already sufficient for reconstruction.

Finally, several artists asked whether our pipeline could provide deformation-related hints (e.g., suggested deformation regions, motion curves, or timing templates), as well as simple lighting-related cues to accelerate the authoring of expressive motion; we do not address these aspects in the current work, but we view them as promising directions for future research built on top of our decomposition.

\section{Additional Comparison for Layer Decomposition}
We provide additional qualitative comparisons with representative layer decomposition baselines, including Qwen-Image-Layered~\cite{qwen-image-layered} and SAM3~\cite{sam3}. We include more examples across diverse character styles and poses, and report the corresponding reconstructed composites to highlight differences in layer quality and usability for downstream 2.5D animation.

\section{Additional Results on Layer Decomposition and Drawing-order Inference}

We present additional qualitative results on layer decomposition and drawing-order inference in the following figures (Figures~\ref{fig:supp_qualitative_1}--\ref{fig:supp_qualitative_7}), showing near-perfect layer decomposition and reconstruction. Overall, the framework produces clean and coherent layers that are directly usable for 2.5D animation; however, a small number of examples contain minor artefacts. We include these cases not as failures in practice: they are typically easy to correct in standard layer editors, but as an opportunity to better understand the remaining ambiguities in our current pipeline.

One representative example, Figure~\ref{fig:supp_qualitative_5}, shows an overlap between \textit{Topwear} and \textit{Bottomwear}. While such overlaps can be easily removed with minimal manual edits (e.g., in Photoshop) based on the visibility mask, the extra content can also be useful in animation, since it effectively provides an additional under-layer for dress motion. We believe this behaviour is largely caused by semantic ambiguity during labelling: for dresses, the boundary between top and bottom garments is not always well-defined, and our dataset can contain locally ambiguous supervision.

Another example, Figure~\ref{fig:supp_qualitative_6}, achieves an almost identical reconstruction, but places the shoe tongue slightly in front of the leg. Interestingly, the model still hallucinates the occluded shoe geometry plausibly to provide reasonable and visually pleasing reconstruction.
We hypothesise that this artefact primarily stems from insufficient stratification for \textit{Footwear}. In most cases footwear does not require interleaving, so we do not stratify it by default; adding a footwear-specific stratification rule would likely resolve this scenario.

Finally, we include an example with dense hand-drawn hatching (Figure~\ref{fig:supp_qualitative_7}). The framework produces a reasonable decomposition and occlusion completion for this abstract, high-frequency style, but our transparency decoder can attenuate some of the finest strokes, causing partial loss of hatching texture. Improving the preservation of such high-frequency details may require a higher-capacity transparency decoder or a backbone that better retains pixel-level structure, which we leave for future work.

\section{Additional Evaluation of the Body Part Consistency Module}
We provide further ablations of the Body Part Consistency Module in the supplementary material to better characterise its contribution to global completeness and cross-layer coherence. In addition to the quantitative metrics reported in the main paper, we include additional qualitative results that highlight typical failure modes without the module (e.g., missing or under-extracted parts in visually ambiguous regions) and the corresponding improvements when enabling cross-part attention.

\begin{figure*}[t]
    \centering
    \includegraphics[width=\linewidth]{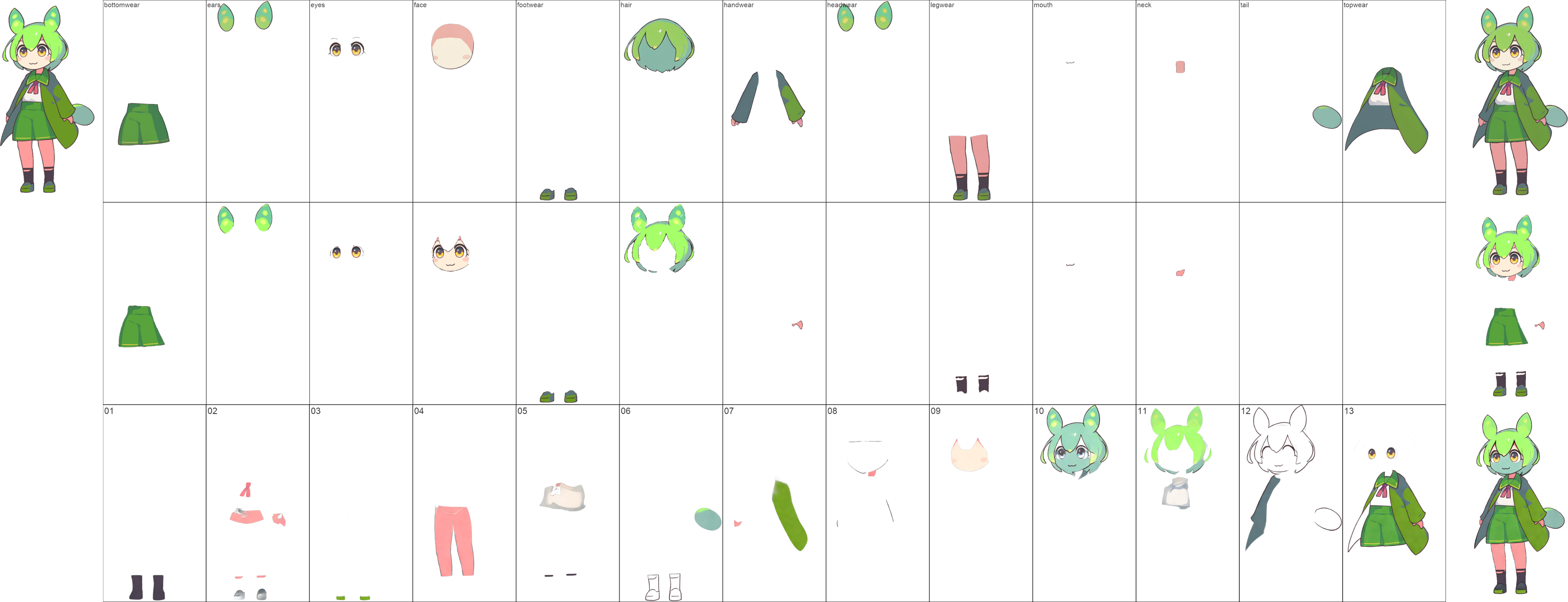}
    \caption{Visual comparison with layer decomposition baselines. Top-to-bottom: Ours (2.5D), SAM3 (2D), Qwen-Image-Layered (2.5D). The rightmost column shows the reconstruction.}
    \label{fig:supp_baseline_1}
\end{figure*}

 \begin{figure*}[t]
    \centering
    \includegraphics[width=\linewidth]{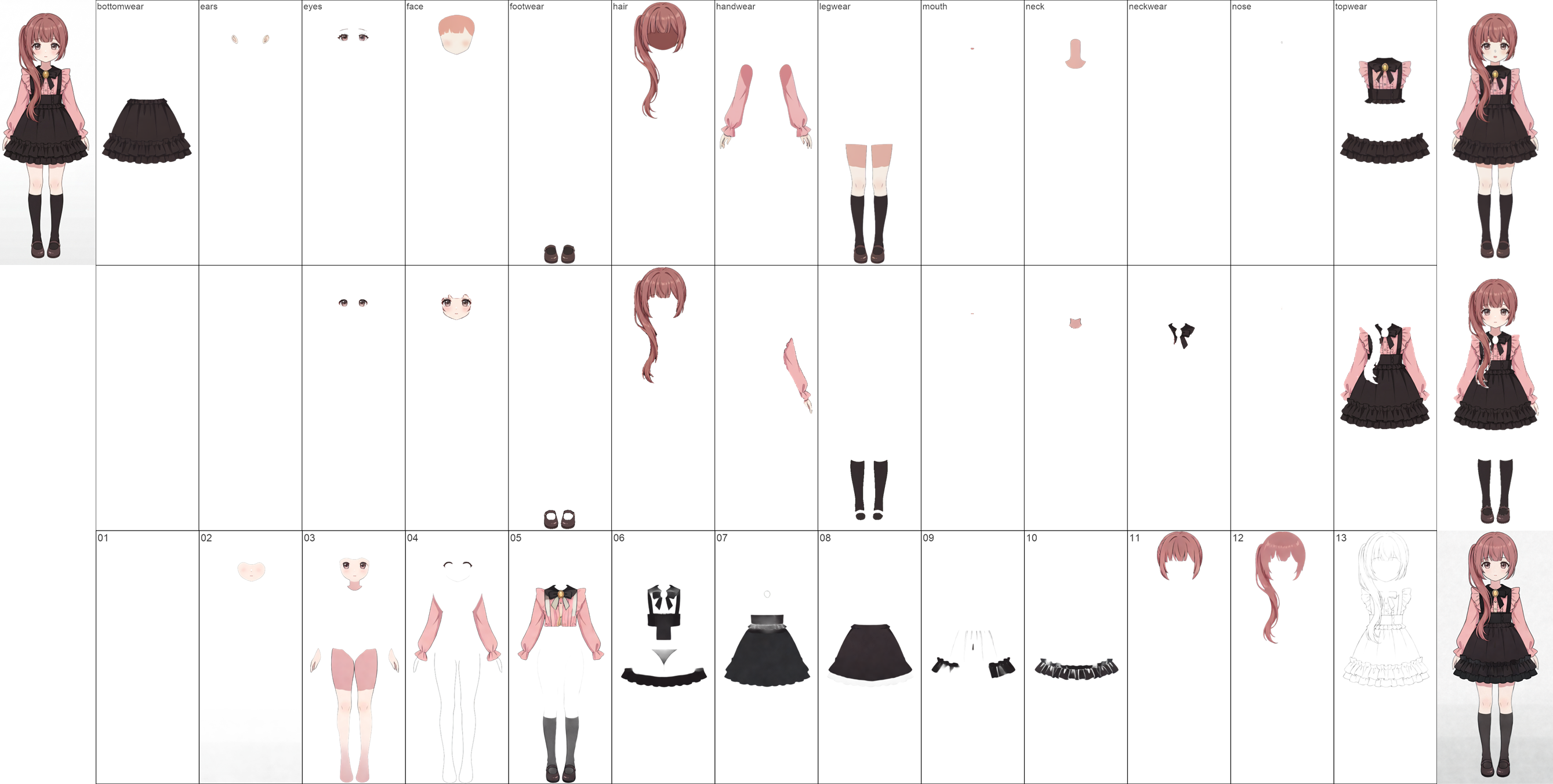}
    \caption{Visual comparison with layer decomposition baselines. Top-to-bottom: Ours (2.5D), SAM3 (2D), Qwen-Image-Layered (2.5D). The rightmost column shows the reconstruction.}
    \label{fig:supp_baseline_2}
\end{figure*}

\begin{figure*}[t]
    \centering
    \includegraphics[width=\linewidth]{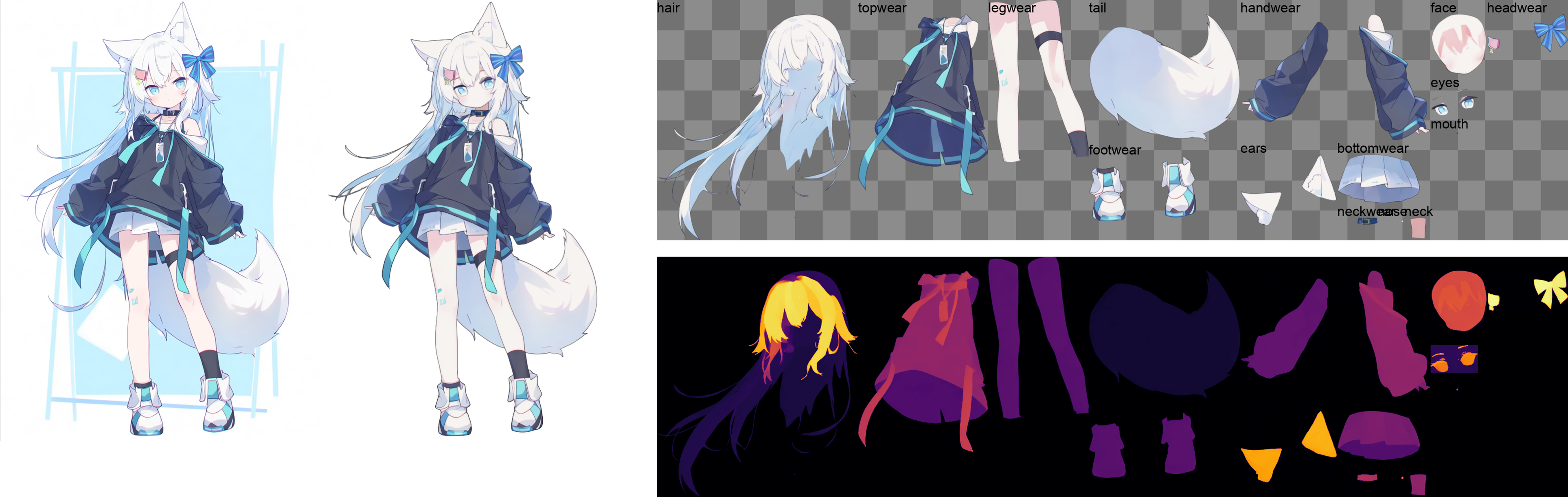}
    \caption{Visualisation of layer decomposition and drawing-order inference with our proposed framework. We also demonstrate the reconstruction.}
    \label{fig:supp_qualitative_1}
\end{figure*}

\begin{figure*}[t]
    \centering
    \includegraphics[width=\linewidth]{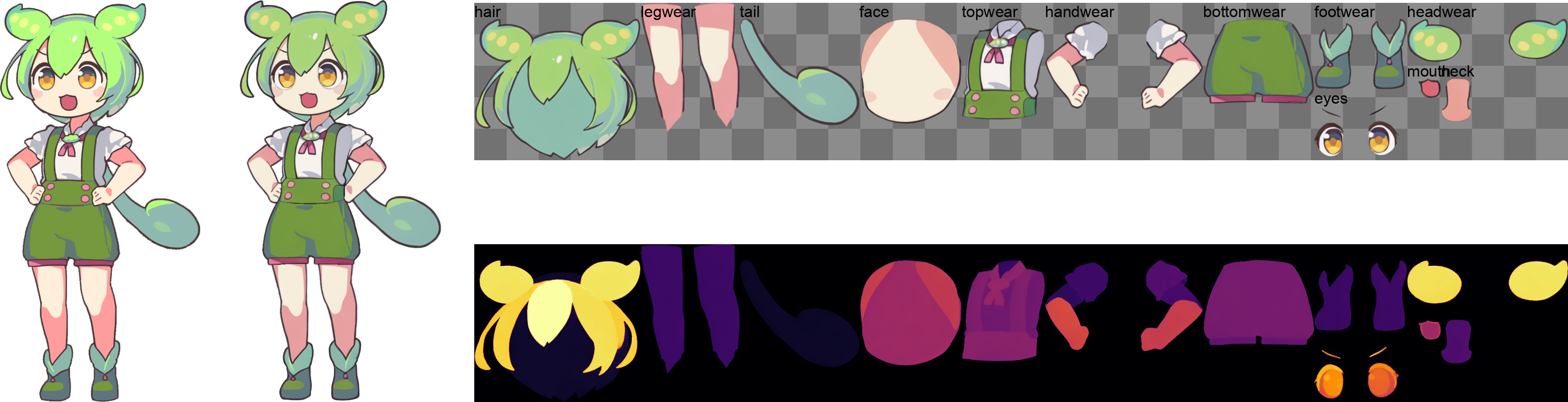}
    \caption{Visualisation of layer decomposition and drawing-order inference with our proposed framework. We also demonstrate the reconstruction.}
    \label{fig:supp_qualitative_2}
\end{figure*}

\begin{figure*}[t]
    \centering
    \includegraphics[width=\linewidth]{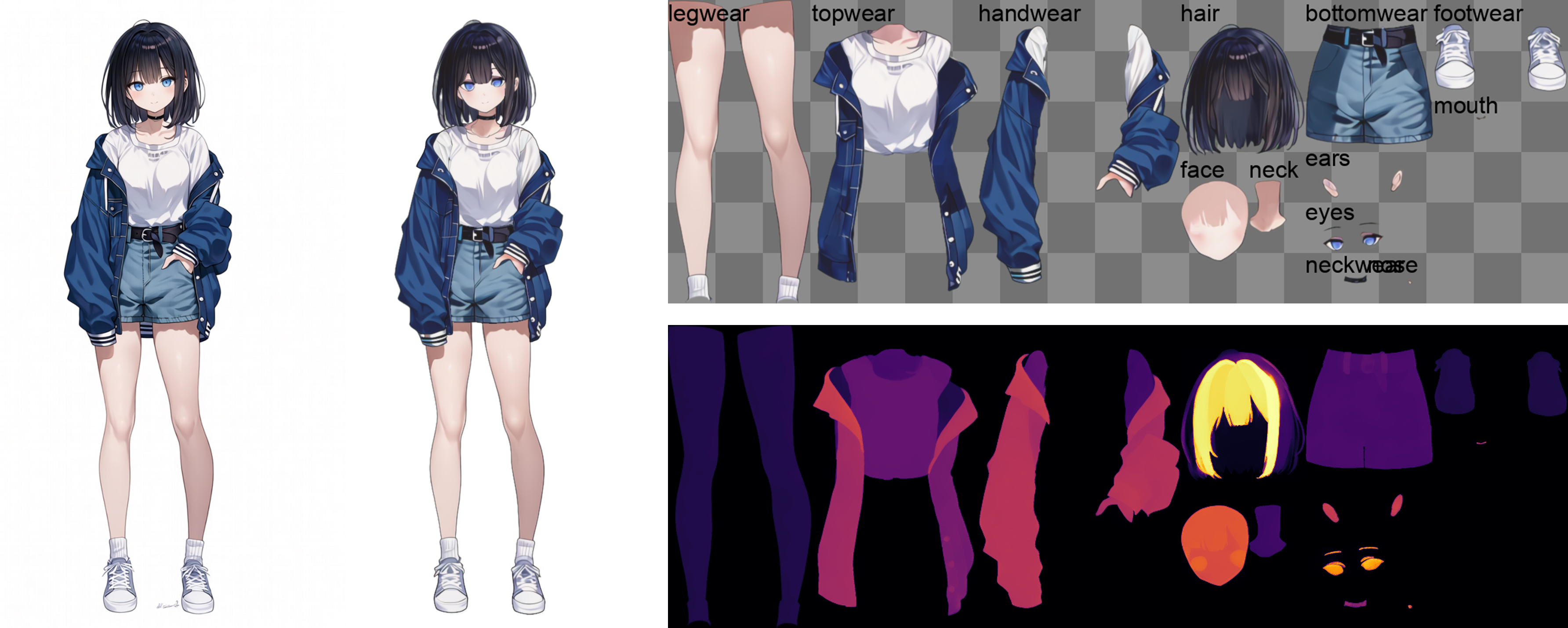}
    \caption{Visualisation of layer decomposition and drawing-order inference with our proposed framework. We also demonstrate the reconstruction.}
    \label{fig:supp_qualitative_3}
\end{figure*}

\begin{figure*}[t]
    \centering
    \includegraphics[width=\linewidth]{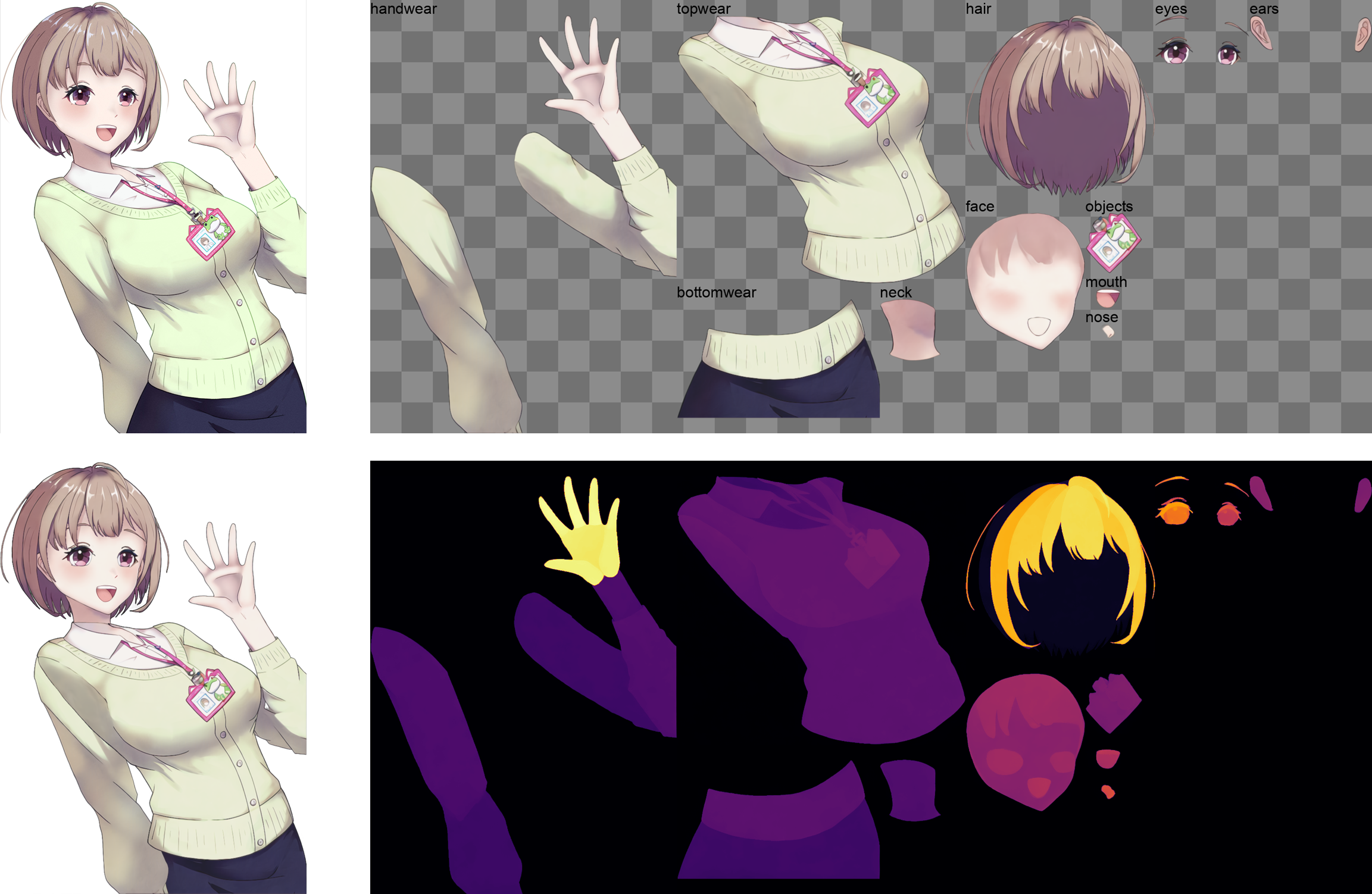}
    \caption{Visualisation of layer decomposition and drawing-order inference with our proposed framework. We also demonstrate the reconstruction.}
    \label{fig:supp_qualitative_4}
\end{figure*}

\begin{figure*}[t]
    \centering
    \includegraphics[width=\linewidth]{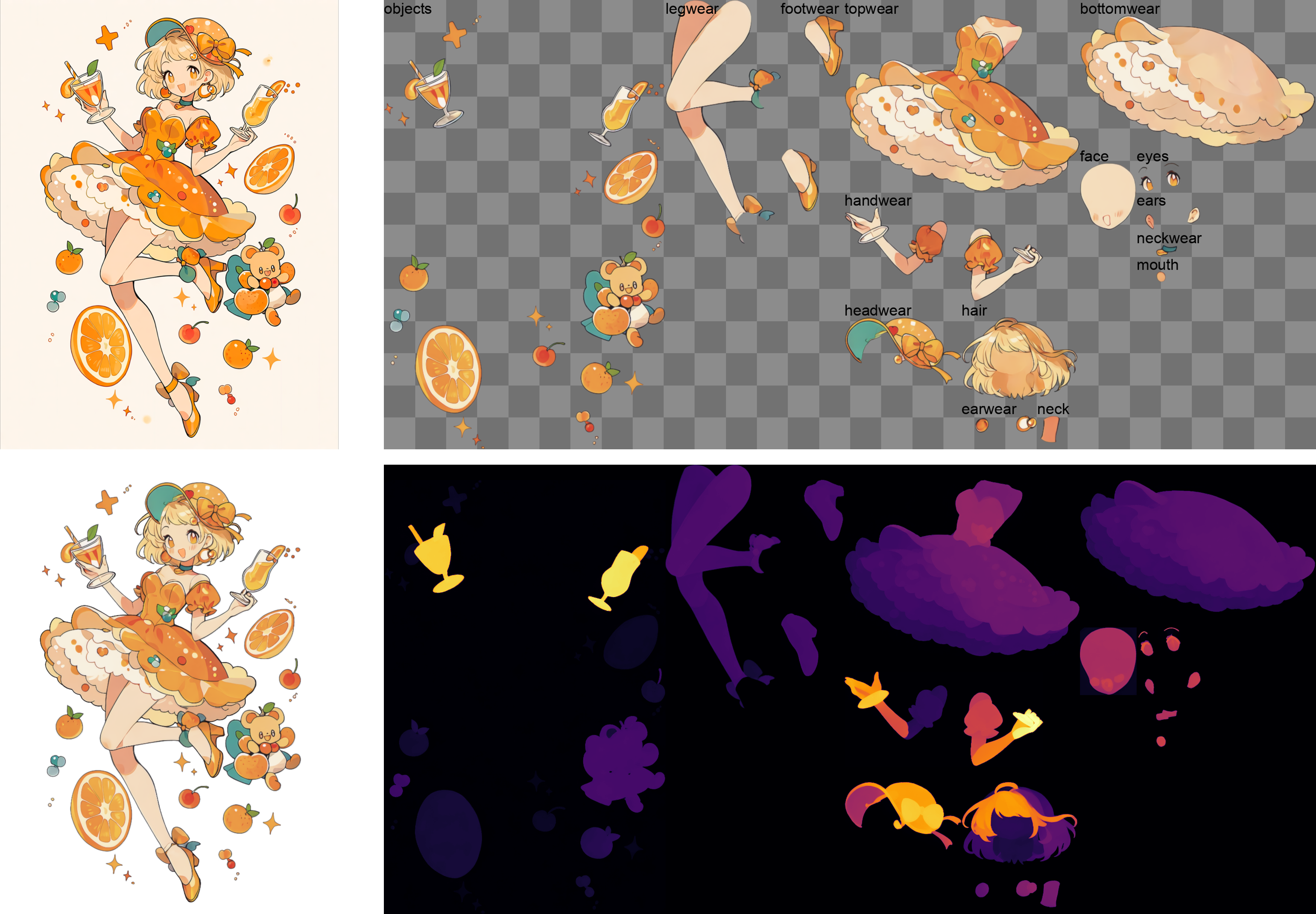}
    \caption{Visualisation of layer decomposition and drawing-order inference with our proposed framework. We also demonstrate the reconstruction.}
    \label{fig:supp_qualitative_5}
\end{figure*}

\begin{figure*}[t]
    \centering
    \includegraphics[width=\linewidth]{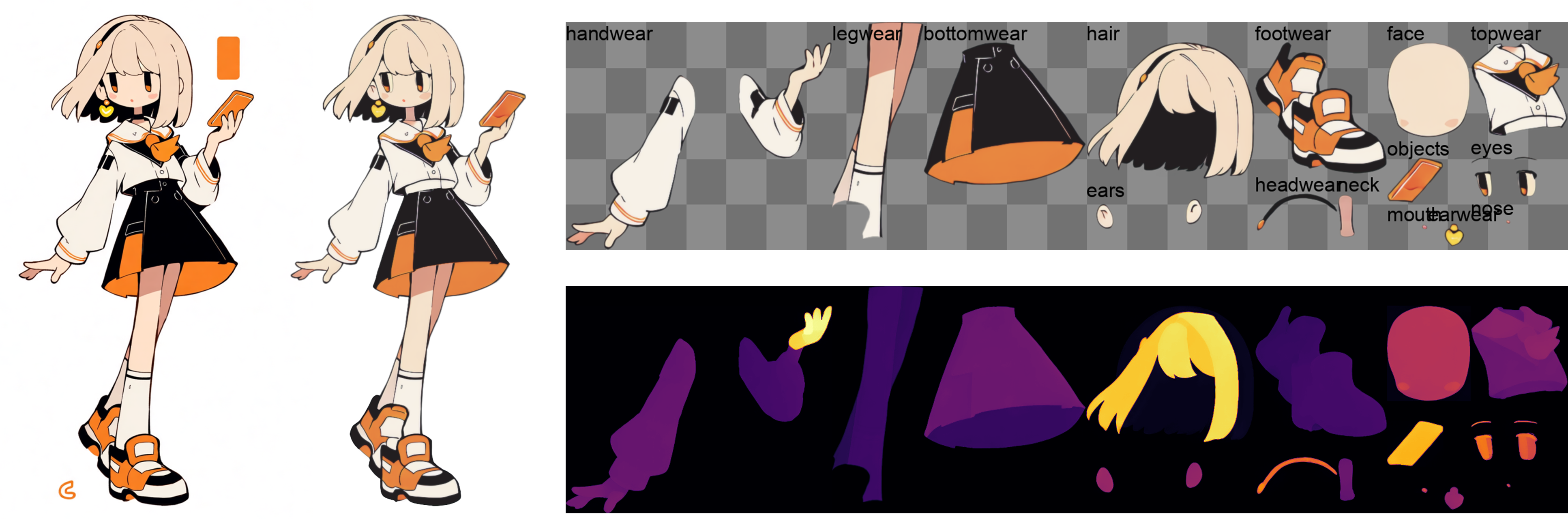}
    \caption{Visualisation of layer decomposition and drawing-order inference with our proposed framework. We also demonstrate the reconstruction.}
    \label{fig:supp_qualitative_6}
\end{figure*}

\begin{figure*}[t]
    \centering
    \includegraphics[width=\linewidth]{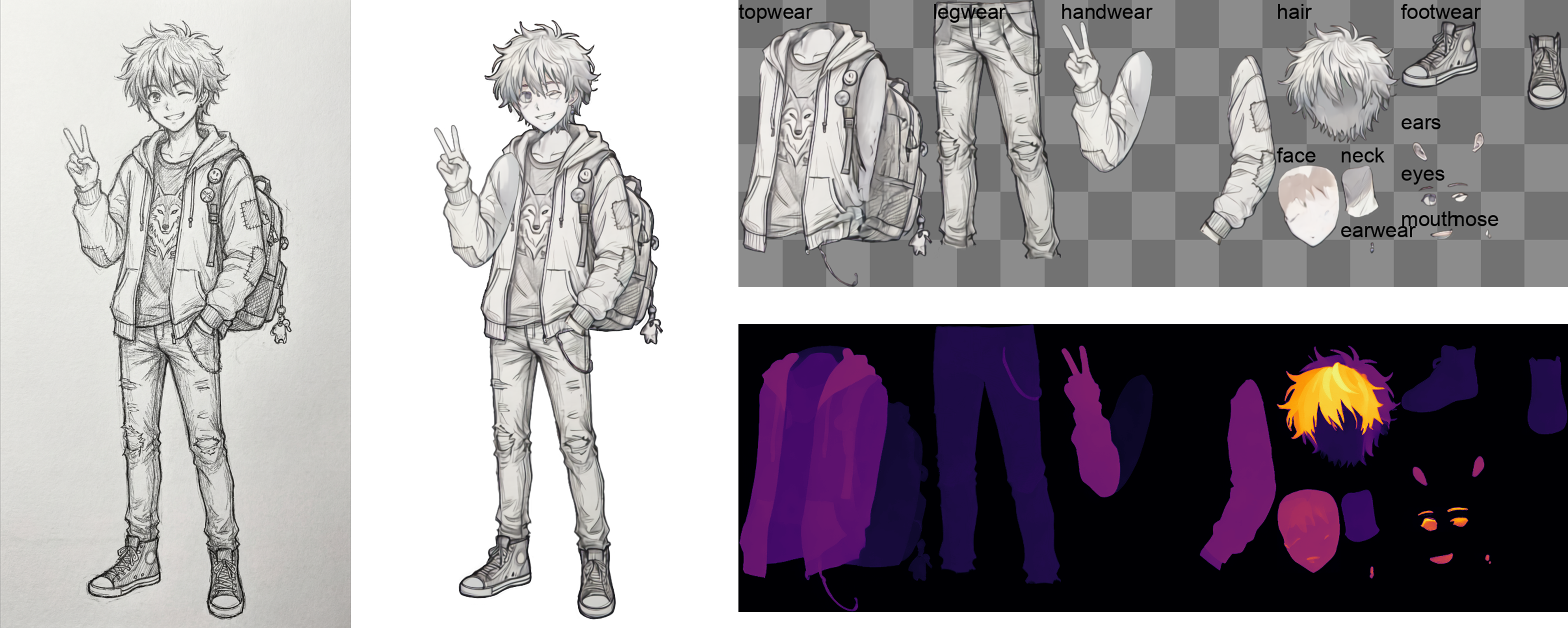}
    \caption{Visualisation of layer decomposition and drawing-order inference with our proposed framework. We also demonstrate the reconstruction.}
    \label{fig:supp_qualitative_7}
\end{figure*}

 \begin{figure*}[t]
    \centering
    \includegraphics[width=\linewidth]{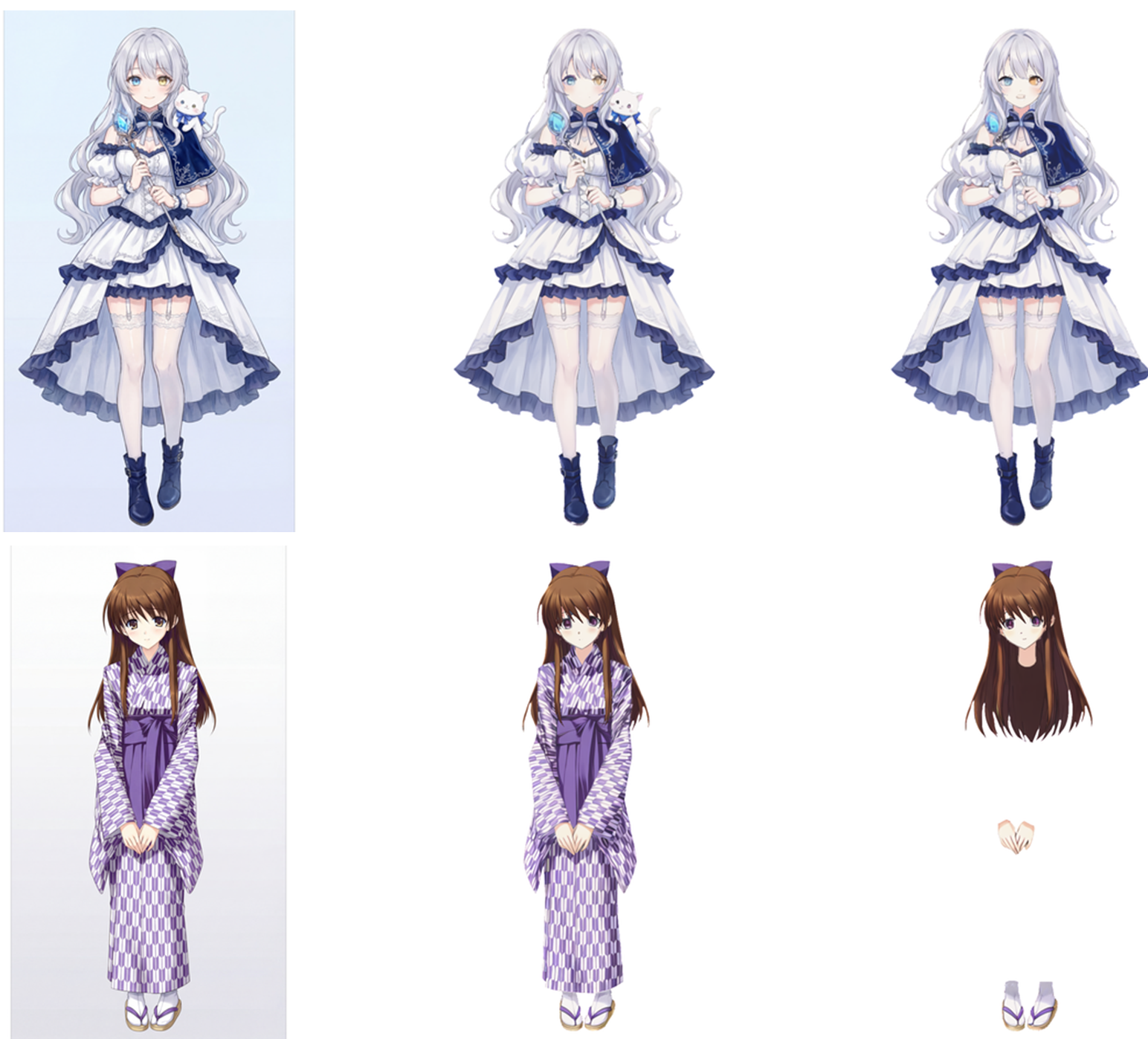}
    \caption{Additional ablation results for the Body Part Consistency Module. From left to right: Input, Ours (Full), Ours (without the Module). Without the module, the model often leaves some parts incomplete or allocates ambiguous regions to neighbouring layers, causing errors in reconstruction.}
\end{figure*}

\end{document}